%% file: main.tex
\documentclass[10pt,twocolumn,letterpaper]{article}

\usepackage[pagenumbers]{cvpr} % To force page numbers, e.g. for an arXiv version

\input{preamble}

\definecolor{cvprblue}{rgb}{0.21,0.49,0.74}
\usepackage[pagebackref,breaklinks,colorlinks,allcolors=cvprblue]{hyperref}
\usepackage[table]{xcolor}
\usepackage{booktabs}
\usepackage{multirow}

\definecolor{panelYellow}{RGB}{250,249,228}
\def\confName{CVPR}
\def\confYear{2027}

\title{Beyond Image–Text Pairs: Aligning Semantic Changes for\\ Fine-Grained Visual Understanding}

\author{
Yujie Sun$^{1\dagger}$,
Zeqing Wang$^{2\dagger}$,
Shuang Liang$^{3\dagger}$,
Zhenlin Zhao$^{4,5}$,
Kunyu Zhang$^{6}$,
Xiaowu Sun$^{1,7*}$ \\[4pt]
{\normalsize
$^{1}$\,Xi'an Jiaotong-Liverpool University \quad
$^{2}$\,Sun Yat-sen University \quad
$^{3}$\,The University of Hong Kong
}\\[2pt]
{\normalsize
$^{4}$\,Cleer Science Co., Ltd. \quad
$^{5}$\,Xi'an Jiaotong University \quad
$^{6}$\,Zhengzhou University
}\\[2pt]
{\normalsize
$^{7}$\,\'Ecole polytechnique f\'ed\'erale de Lausanne (EPFL)
}\\[4pt]
{\small\texttt{xiaowu.sun@xjtlu.edu.cn}}\\[2pt]
{\small\url{https://06yujie.github.io/SDA/}}
}

\makeatletter
\g@addto@macro\@maketitle{%
    \par
    \vspace{-18pt}
    \input{figures/banner}
}
\makeatother
\begin{document}

\maketitle
\begingroup
\renewcommand{\thefootnote}{\fnsymbol{footnote}}
\footnotetext[2]{Equal contribution.\quad
$^{*}$Corresponding author.}
\endgroup
\input{sec/0_abstract}    
\input{sec/1_intro}
\input{sec/2_related}
\input{sec/3_method}
\input{sec/4_experiments}
\input{sec/5_conclusion}

% \clearpage
% \section*{LLM Usage Statement}
% ChatGPT (OpenAI) was used to assist with manuscript preparation, including language refinement, grammatical corrections, and improvements to the clarity, coherence, and overall organization of the text. All AI-assisted content was carefully reviewed, verified, and revised by the authors, who take full responsibility for the accuracy, integrity, and conclusions of this work.

{
    \small
     \bibliographystyle{ieeenat_fullname}
     \bibliography{main}
}

\input{sec/X_suppl}

\end{document}

%% file: preamble.tex
\usepackage{booktabs}
\usepackage{multirow}
\usepackage{makecell}
\usepackage{graphicx}
\usepackage{float}
\usepackage{cuted}

\usepackage{tcolorbox}
\usepackage{listings}

\tcbset{varfamilyprompt/.style={
  colframe=blue!20!black,
  colback=gray!10,
  coltitle=black,
  colbacktitle=blue!20,
  fonttitle=\small\bfseries,
  fontupper=\small,
  boxrule=0.6pt,
  arc=1.5mm,
  left=6pt,right=6pt,top=4pt,bottom=4pt,
  before skip=4pt,after skip=4pt
}}

\lstdefinestyle{varfamilyprompttext}{
  basicstyle=\small\ttfamily,
  columns=fullflexible,
  keepspaces=true,
  showstringspaces=false,
  breaklines=true,
  breakatwhitespace=false,
  breakindent=0pt,
  aboveskip=3pt,
  belowskip=3pt,
  frame=none,
  numbers=none,
  language={}
}

%% file: figures/banner.tex
\begin{center}
\includegraphics[width=\textwidth]{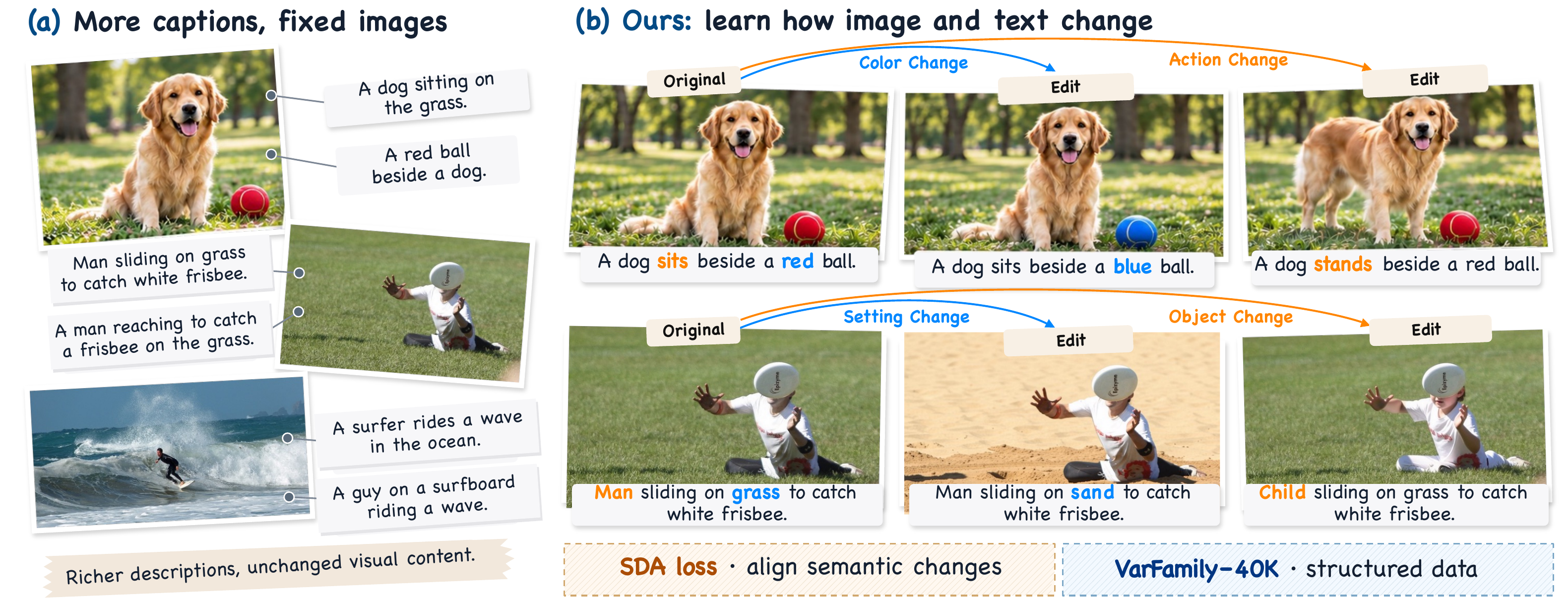}
\begin{minipage}{\textwidth}
\captionsetup{hypcap=false,width=\linewidth}
\vspace{3pt}
\captionof{figure}{
    \textbf{From richer descriptions to aligned semantic changes.}
    (a) Multiple captions describe complementary aspects of a fixed image.
    (b) Our approach constructs image-text variants with corresponding
    semantic changes while preserving content outside the targeted edit.
    These variants provide supervision for SDA loss, which aligns visual and textual changes.
}
\label{fig:banner}
\end{minipage}

\end{center}

%% file: sec/0_abstract.tex
\begin{abstract}
Image--text alignment serves as a foundation for modern vision-language models (VLMs), yet prevailing training paradigms provide limited supervision for learning how semantic changes correspond across modalities. To address this limitation, we propose a post-training method that learns cross-modal semantic change correspondences through controlled visual and textual variations. Specifically, we introduce a \textbf{S}emantic \textbf{D}ifference \textbf{A}lignment (\textbf{SDA}) loss which can align the directions of corresponding visual and textual representation changes. We combine our proposed SDA with pairwise sigmoid matching to jointly learn image–text correspondences and cross-modal semantic changes, encouraging more fine-grained and discriminative visual representations. To support this learning, we construct VarFamily-40K, a dataset of image--text pairs with controlled cross-modal semantic variations. Experiments demonstrate consistent improvements in compositional understanding and downstream classification performance, alongside further gains on multimodal tasks when the adapted visual encoder is integrated into VLMs.
\end{abstract}

%% file: sec/1_intro.tex
\section{Introduction}
\label{sec:intro}

% Learning semantic correspondences between images and natural language is a central objective of vision-language model pretraining~\cite{radford2021clip,jia2021align,zhai2023siglip}. Specifically, image--text matching approaches such as SigLIP~\cite{zhai2023siglip} associate each positive pair with an image and a description of its content. Howerver, an image naturally can contain multiple objects, attributes, actions, and relationships, while a single description typically captures only a subset of them. For example, as shown in Fig.~\ref{fig:banner}(a), the caption ``A dog sitting on the grass'' captures the dog's posture and setting, while leaving other visible details unspecified in the description, such as the ``red ball'' in the image. Thus, each image--text correspondence in current cross-model supervision conveys only a partial view of the scene, motivating richer and more detailed forms of cross-modal supervision.

With the development of VLMs~\cite{wang2026towards,bai2025qwen25vl,chen2024internvl25,yao2024minicpmv,liu2024llavanext}, learning semantic correspondences between images and natural language is a central objective of VLM pretraining~\cite{radford2021clip,jia2021align,zhai2023siglip}. Existing approaches, such as CLIP~\cite{radford2021clip} and SigLIP~\cite{zhai2023siglip}, learn these correspondences by matching images with their associated textual descriptions. However, an image naturally contains multiple objects, attributes, actions, and relationships, while a single description typically captures only a subset of them, which limits the performance of current VLMs~\cite{wang2025generated,wang2026phydetex,wang2026timecausality}. A natural approach to enriching image--text supervision is therefore to associate each image with multiple captions, allowing complementary descriptions to capture a broader range of visual semantics~\cite{lai2024veclip}. For example, as shown in Fig.~\ref{fig:banner}(a), the captions ``A dog sitting on the grass'' and ``A red ball beside a dog'' describe the dog's posture and setting, and the ball's color and spatial relationship to the dog, respectively.

% A natural approach to enriching image--text supervision from such a view is to associate each image with multiple captions, allowing complementary descriptions to capture a broader range of visual semantics~\cite{lai2024veclip}. Existing work has explored sampling sub-captions from detailed descriptions~\cite{zheng2024dreamlip}, combining synthetic captions with original web text~\cite{lai2025revisit}, and distributing supervision across sentences in long captions~\cite{lavoie2026shortsighted}. However, increasing textual diversity does not explicitly establish how semantic changes and what the semantic meaning in language correspond to changes in visual content. Since these descriptions refer to the same image, their semantic variations are not accompanied by corresponding changes in the visual modality. This distinction is particularly relevant to fine-grained visual understanding, where subtle variations in attributes, actions, or relationships can alter the semantic correspondence between otherwise similar images and texts~\cite{thrush2022winoground,hsieh2023sugarcrepe}. We therefore seek to explicitly model corresponding semantic changes across the visual and textual modalities.

Existing work has explored sampling sub-captions from detailed descriptions~\cite{zheng2024dreamlip}, combining synthetic captions with original web text~\cite{lai2025revisit}, and distributing supervision across sentences in long captions~\cite{lavoie2026shortsighted}. These approaches enrich textual supervision by describing complementary aspects of a fixed image, while leaving the corresponding visual content unchanged. Although image--text matching can learn to distinguish fine-grained semantics, its supervision does not directly constrain how a change in visual representation relates to the corresponding change in textual representation. This distinction is particularly relevant to fine-grained visual understanding, where subtle variations in attributes, actions, or relationships can alter the semantic correspondence between otherwise similar images and texts~\cite{thrush2022winoground}. We therefore seek to explicitly model corresponding semantic changes across the visual and textual modalities.

% To this end, we first introduce \textbf{S}emantic \textbf{D}ifference \textbf{A}lignment (\textbf{SDA}) loss, which makes the correspondence between visual and textual changes an explicit training target during the alignment of visual and textual modalities. Pairwise sigmoid matching supervises whether individual images and captions match, but does not directly constrain how the visual difference between two images relates to the semantic difference between their descriptions. By contrast, our SDA loss captures this additional relationship: as illustrated in Fig.~\ref{fig:banner}(b), a change from a red ball to a blue ball should induce a visual embedding difference aligned with the corresponding textual difference. A directional alignment loss establishes this alignment, while a margin-based discrimination loss encourages each visual change to align more closely with its paired textual change than with textual changes from other image--text variant groups.

Our approach builds on the intuition that corresponding semantic changes in images and texts can be represented by differences between their original and edited embeddings. Since both changes express the same semantic transition, their embedding differences can be explicitly aligned in a shared space. To this end, we introduce \textbf{S}emantic \textbf{D}ifference \textbf{A}lignment (\textbf{SDA}) loss, which makes the correspondence between visual and textual changes an explicit training target. As illustrated in Fig.~\ref{fig:banner}(b), a change from a red ball to a blue ball should induce a visual embedding difference aligned with the corresponding textual difference. A directional alignment loss encourages corresponding visual and textual embedding differences to follow similar directions, while a margin-based discrimination loss encourages each visual change to align more closely with its paired textual change than with textual changes from other image--text variant groups. We combine SDA loss with pairwise sigmoid matching to supervise both individual image--text correspondences and corresponding semantic changes.

% Though our SDA loss models the detailed relationship between visual and textual changes, training with SDA loss requires corresponding visual and textual changes, which multiple captions of a fixed image do not directly provide. To provide this supervision, we further construct VarFamily-40K, a dataset of image--text variant families linking original pairs to semantically edited counterparts. Each family is created through a targeted caption modification and a corresponding image edit, followed by semantic consistency filtering. Preserving scene content outside the targeted edit helps isolate the visual change specified by the text. For example, an edit that changes a ball's color from red to blue should preserve the dog's posture and the surrounding scene. Learning from these controlled variations with the joint objective encourages visual representations that distinguish fine-grained semantic differences. Overall, our main contributions are summarized as follows:

Training with SDA loss requires image--text pairs with corresponding and controlled semantic changes. However, existing multi-caption datasets associate different descriptions with a fixed image and therefore do not directly provide this supervision. To address this limitation, we construct \textbf{VarFamily-40K}, a dataset of image--text variant families linking original pairs to semantically edited counterparts. Each family is created through a targeted caption modification and a corresponding image edit, followed by semantic consistency filtering. Preserving scene content outside the targeted edit helps isolate the visual change specified by the text. For example, the ball-color edit in Fig.~\ref{fig:banner}(b) should preserve the dog's posture and the surrounding scene. Together, SDA loss and VarFamily-40K enable post-training that encourages visual representations to distinguish fine-grained semantic differences. Our main contributions are summarized as follows:

\begin{itemize}
    \item We propose SDA loss to explicitly align semantic changes across visual and textual representations, enabling more fine-grained cross-modal supervision.

    \item We construct VarFamily-40K, a dataset of controlled image--text variant families that provides semantic changes for SDA-based post-training.

    \item Extensive experiments demonstrate that, combined with VarFamily-40K, our SDA loss improves both the downstream classification performance and general visual understanding in VLMs across ten multimodal benchmarks.

\end{itemize}

%% file: sec/2_related.tex
\section{Related Work}

%\subsection{Image--Text Pretraining Objectives}
\noindent\textbf{Image--Text Pretraining Objectives.}
Existing image--text pretraining objectives primarily learn cross-modal alignment by distinguishing matched image--text pairs from mismatched ones. CLIP~\cite{radford2021clip} and ALIGN~\cite{jia2021align} employ contrastive objectives that normalize matching scores across candidate images or texts within a batch, coupling the optimization of different pairs. SigLIP~\cite{zhai2023siglip} instead adopts a pairwise sigmoid loss that evaluates each image--text pair without global softmax normalization, achieving better performance at smaller batch sizes than softmax-based contrastive learning. Both objectives assign a score to each image–text pair. When a caption changes from a red ball to a blue ball, the scores indicate which image matches each caption. The training objective leaves open how the displacement between the two image embeddings relates to the displacement between the caption embeddings. Our post-training method retains pairwise sigmoid matching as the base objective and introduces SDA loss to additionally supervise corresponding semantic changes across modalities.

%\subsection{Fine-Grained Vision--Language Alignment}
\noindent\textbf{Fine-Grained Vision--Language Alignment.}
Recent advances in fine-grained vision–language alignment have focused on enriching textual supervision and learning more precise correspondences between visual content and language. LaCLIP~\cite{fan2023laclip} uses large language model (LLM)-generated caption rewrites to increase linguistic diversity, while Long-CLIP~\cite{zhang2024longclip} extends CLIP to accommodate longer descriptions containing richer visual details. LLIP~\cite{lavoie2024llip} learns multiple visual features and combines them based on the accompanying text, allowing different captions to emphasize different aspects of the same image. Other approaches refine alignment at the region or patch level. RegionCLIP~\cite{zhong2022regionclip} aligns image regions with textual concepts, while FILIP~\cite{yao2022filip} and SPARC~\cite{bica2024sparc} learn correspondences between image patches and text tokens. DreamLIP~\cite{zheng2024dreamlip} connects sampled subcaptions with relevant image patches, and FG-CLIP 2~\cite{xie2026fgclip2} combines detailed captions, region-level alignment, and discriminative supervision. Together, these methods enrich semantic coverage and refine cross-modal correspondences. Their supervision focuses on associating language with existing visual content, without explicitly aligning the representation differences induced by corresponding semantic changes in images and texts.

%\subsection{Learning from Semantic Changes}
\noindent\textbf{Learning from Semantic Changes.}
Semantic changes have been explored in vision-language research to improve compositional understanding and support image editing. For example, NegCLIP~\cite{yuksekgonul2023when} introduces composition-aware hard negatives through textual perturbations, while TripletCLIP~\cite{patel2024tripletclip} further incorporates synthetic visual negatives to improve compositional discrimination. Some other methods, like Instruct-CLIP~\cite {chen2025instructclip}, learn to align representations of changes between original and edited images with instructions, supporting instruction refinement and instruction-guided image editing. Our method instead aligns the embedding differences between original and edited images with those between their corresponding captions, using these paired differences for training.

%% file: sec/3_method.tex
\section{Method}
\label{sec:method}

\begin{figure*}[t]
    \centering
    \includegraphics[width=\textwidth]{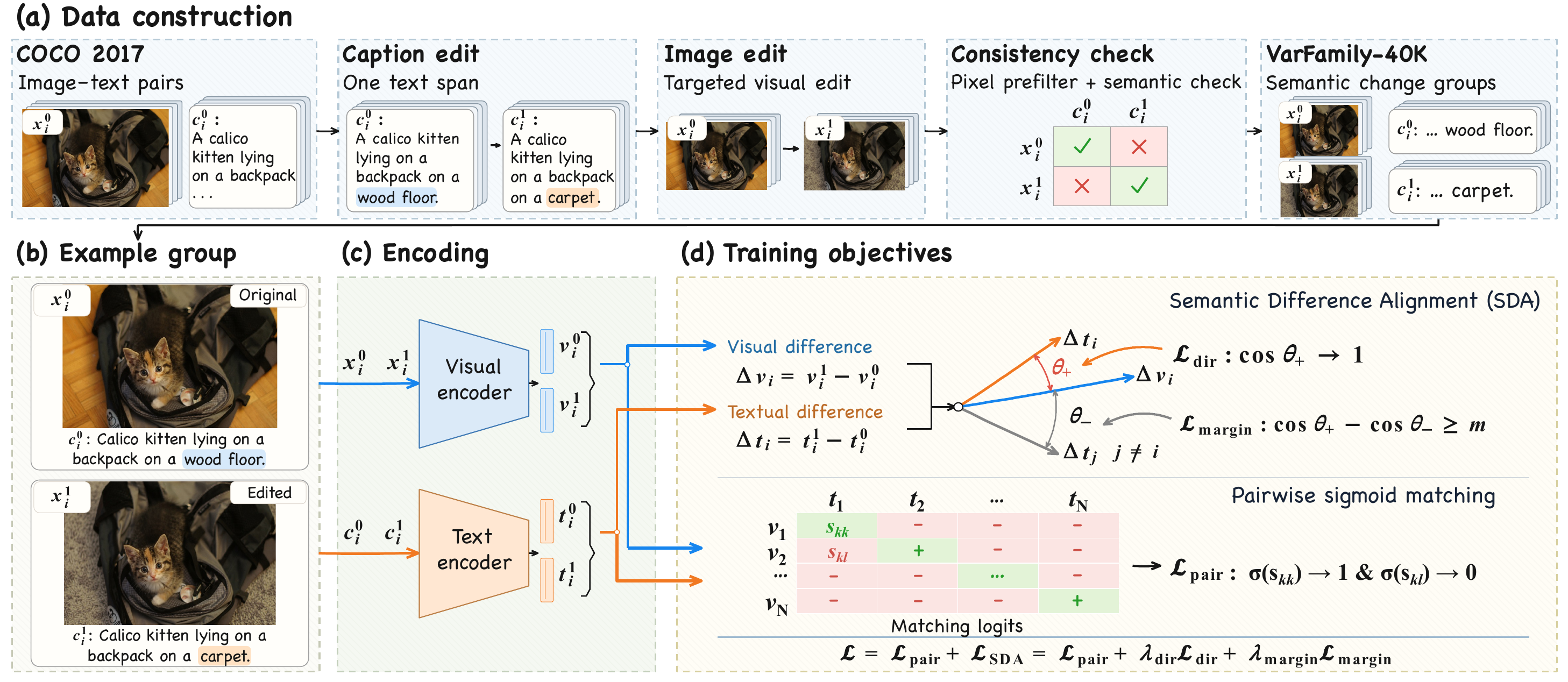}
    \captionsetup{skip=4pt}
    \caption{
    Overview of our dataset construction and proposed post-training method. We construct VarFamily-40K with controlled semantic changes across image--text pairs and jointly optimize SDA loss and pairwise sigmoid matching to improve visual representations.
}
    \vspace{-3mm}
    \label{fig:overview}
\end{figure*}

Our method aims to improve visual representations by learning how semantic changes correspond across images and texts. Rather than relying solely on individual image--text correspondences, we introduce SDA loss to align corresponding visual and textual changes while distinguishing them from changes in other groups. To support this, we construct VarFamily-40K, which provides controlled semantic variations through original and edited image--text pairs. Fig.~\ref{fig:overview} provides an overview of the proposed approach.

\subsection{Semantic Difference Alignment Loss}
\label{sec:difference_alignment}

We consider an image–text group $\mathcal{G}_i = \{(x_i^0,c_i^0),(x_i^1,c_i^1)\}$, where $x_i^a$ and $c_i^a$ denote the image and its corresponding caption, respectively, with $a=0$ indicating the original pair and $a=1$ its semantically edited counterpart. The two pairs share scene context and differ in a targeted semantic concept. A training batch consists of $B$ such groups, containing $2B$ images and $2B$ captions in total.

For each image–text group, we extract visual and textual representations using an image encoder $f_\theta$ and a text encoder $g_\phi$, respectively, obtaining $\ell_2$-normalized embeddings $\mathbf{v}_i^a = f_\theta(x_i^a)$ and $\mathbf{t}_i^a = g_\phi(c_i^a)$. To characterize the targeted semantic modification, we define the visual and textual difference vectors between the original and edited representations as
\begin{equation}
\vspace{-2mm}
\Delta\mathbf{v}_i = \mathbf{v}_i^1 - \mathbf{v}_i^0, \qquad
\Delta\mathbf{t}_i = \mathbf{t}_i^1 - \mathbf{t}_i^0.
\vspace{-1mm}
\end{equation}

Each difference vector describes a shift from the original to the edited representation, with its direction indicating how the representation changes in the embedding space. Since the visual and textual differences correspond to the same semantic modification, we encourage their directions to be consistent across modalities. To measure this consistency, we compute the cosine similarity between visual and textual difference vectors:
\begin{equation}
    D_{ij} =
    \Delta\mathbf{v}_i^\top\Delta\mathbf{t}_j
    \big/
    \bigl(\|\Delta\mathbf{v}_i\|_2\,\|\Delta\mathbf{t}_j\|_2\bigr).
\end{equation}
Here, $D_{ii}$ measures the directional similarity between corresponding visual and textual changes, whereas $D_{ij}$ ($i\neq j$) measures the similarity between changes from different groups. We define the directional alignment loss $\mathcal{L}_{\mathrm{dir}}$ across the $B$ groups as
\begin{equation}
    \vspace{-2mm}
    \mathcal{L}_{\mathrm{dir}}=
    \frac{1}{B}\sum_{i=1}^{B}(1-D_{ii}).
    \label{eq:direction_loss}
    % \vspace{-1mm}
\end{equation}
Although directional alignment encourages corresponding visual and textual differences to follow similar directions, it does not explicitly account for their similarity to differences from other groups. Consequently, a visual difference vector may also exhibit high similarity to non-corresponding textual differences, making it difficult to distinguish the intended semantic change. To address this issue, we encourage the similarity of corresponding differences to exceed that of non-corresponding differences by a specified margin using the following loss $\mathcal{L}_{\mathrm{margin}}$:
% \begin{equation}
%     \vspace{-2mm}
%     \mathcal{L}_{\mathrm{margin}}=
%     \frac{1}{B(B-1)}
%     \sum_{i=1}^{B}\sum_{j\ne i}
%     \max\!\left(0,\,m-D_{ii}+D_{ij}\right),
%     \label{eq:margin_loss}
%     % \vspace{-1mm}
% \end{equation}
\begin{equation}
\mathcal{L}_{\mathrm{margin}}
=\frac{1}{B(B-1)}
\sum_{i=1}^{B}\sum_{j\ne i}
\max\!\bigl(0,m-D_{ii}+D_{ij}\bigr).
\label{eq:margin_loss}
\end{equation}
where $m>0$ is a margin that specifies the minimum desired difference between the similarities of corresponding and non-corresponding changes.

We define the SDA loss as a weighted sum of the two terms:
\begin{equation}
    \mathcal{L}_{\mathrm{SDA}} =
    \lambda_{\mathrm{dir}}\mathcal{L}_{\mathrm{dir}}
    +\lambda_{\mathrm{margin}}\mathcal{L}_{\mathrm{margin}},
    \label{eq:sda_loss}
\end{equation}
where $\lambda_{\mathrm{dir}}$ and $\lambda_{\mathrm{margin}}$ are weighting coefficients.

\subsection{VarFamily-40K: Dataset Construction}
\label{sec:semantic_groups}

\begin{figure}[t]
    \centering
    \includegraphics[width=\linewidth]{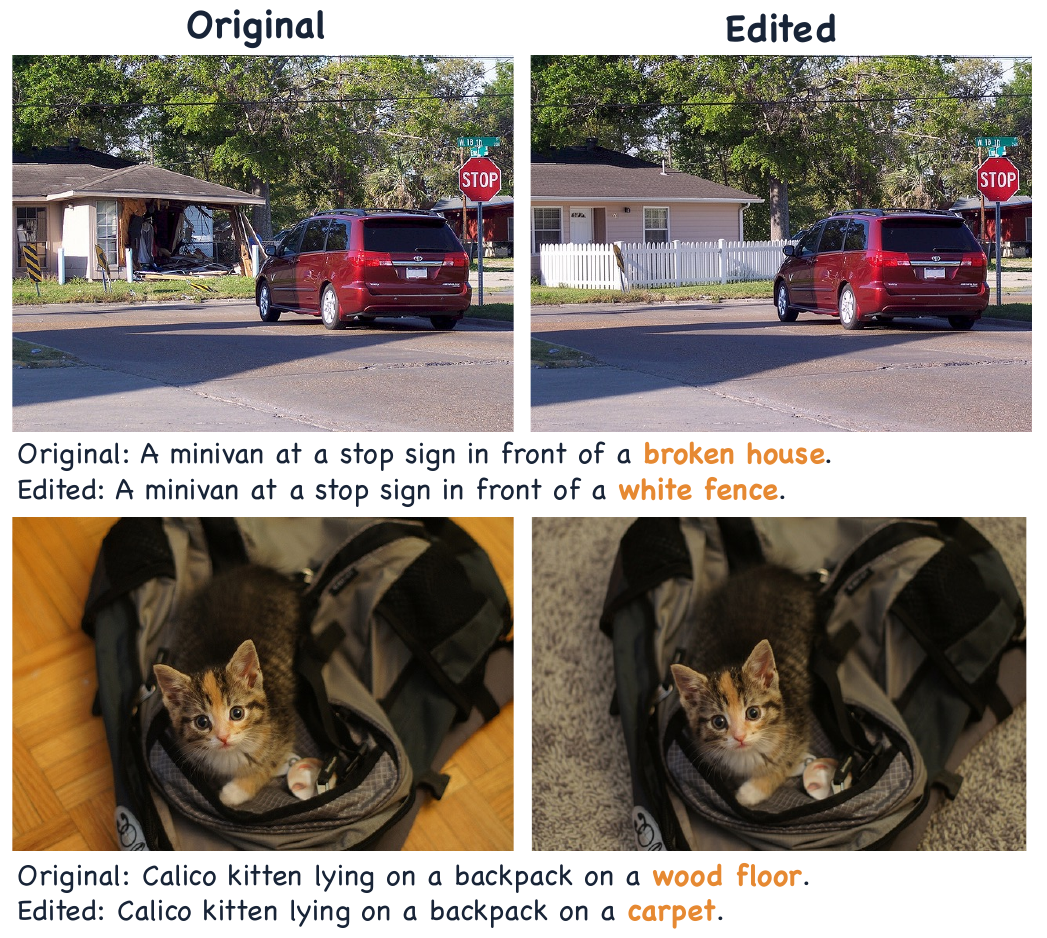}
    \captionsetup{skip=4pt}
    \caption{Examples from VarFamily-40K: original (left) and edited (right) images with corresponding captions.}
    \label{fig:editpairs_examples}
    \vspace{-3mm}
\end{figure}

To provide the corresponding visual and textual changes required by SDA loss, leveraging the progress made by visual content generation methods~\cite{liang2026pixels,wei2026mico,wang2025videoverse}, we construct VarFamily-40K from COCO 2017~\cite{lin2014microsoft}, following the pipeline in Fig.~\ref{fig:overview}(a), which consists of targeted caption rewriting, image editing, and semantic consistency filtering. Each image–text variant group follows the structure defined in Sec.~\ref{sec:difference_alignment}, containing an original image--text pair and its edited counterpart. Fig.~\ref{fig:editpairs_examples} shows example groups from VarFamily-40K.

% \paragraph{(i) Caption rewriting.}
\noindent\textbf{(i) Caption rewriting.}
We use Qwen3.5-9B~\cite{qwen2026qwen35} to generate $c_i^1$ by replacing one short, contiguous span in $c_i^0$, preserving all text outside that span. The modifications cover objects, attributes, actions, quantities, and backgrounds. Each replacement specifies a visually observable semantic change. We validate the rewritten caption against the selected span and apply edit-type and contextual checks. The original and replacement spans are then used to construct an editing prompt that specifies the target change and asks the model to preserve unrelated scene content.

% \paragraph{(ii) Corresponding image editing.}
\noindent\textbf{(ii) Corresponding image editing.}
Based on the obtained editing prompts, we use FLUX.2 Klein 9B~\cite{bfl2025flux2} to generate the edited image $x_i^1$ conditioned on the original image $x_i^0$. We then apply a pixel-difference prefilter to discard generated images with negligible visual changes.

% \paragraph{(iii) Semantic consistency filtering.}
\noindent\textbf{(iii) Semantic consistency filtering.}
Finally, to ensure the quality of our VarFamily-40K, we provide Qwen3-VL-8B-Instruct~\cite{bai2025qwen3vl} with both images and captions and ask it to evaluate each of the four image--text combinations independently. A pair is considered matching when the image clearly supports the full caption, with particular attention to the modified semantic content. We retain groups in which the original and edited images match their respective captions, while neither image matches the other caption. This filtering yields VarFamily-40K, a collection of image–text variant families for post-training. Further details on dataset construction, filtering procedures, and statistics are provided in \textbf{Appendix A}.

\subsection{Post-training with SDA Loss}
\label{sec:joint_optimization}

We adopt a post-training strategy to enhance the fine-grained visual representations of pretrained SigLIP using VarFamily-40K. Specifically, we apply low-rank adaptation (LoRA)~\cite{hu2022lora} to both the visual and textual encoders, keeping the pretrained backbone parameters frozen while updating the LoRA parameters.

While SDA loss aligns semantic changes across visual and textual representations, it does not directly supervise individual image–text correspondences. We therefore retain the original pairwise sigmoid matching objective~\cite{zhai2023siglip} and jointly optimize it with SDA loss to preserve image–text alignment during post-training.

A batch of $B$ groups contains $N=2B$ image–text pairs, with corresponding visual and textual embeddings denoted by $\mathbf{v}_k$ and $\mathbf{t}_k$, respectively, for $k=1,\ldots,N$. The matching image–text pairs serve as positive samples, while all non-corresponding pairs within the mini-batch are treated as negatives. The matching logit between the $k$-th image and the $l$-th caption is computed as
\begin{equation}
    s_{kl}=\alpha\mathbf{v}_k^\top\mathbf{t}_l+\beta,
\end{equation}
where $l=1,\ldots,N$, and $\alpha>0$ and $\beta$ are learnable scale and bias parameters, respectively. We assign a positive label to corresponding image–text pairs and a negative label to non-corresponding pairs, i.e., $y_{kl}=+1$ if $k=l$ and $y_{kl}=-1$ otherwise. The pairwise sigmoid matching loss is then formulated as 
\begin{equation}
    \mathcal{L}_{\mathrm{pair}}=
    -\frac{1}{N^2}
    \sum_{k,l=1}^{N}
    \log\sigma\!\left(y_{kl}s_{kl}\right),
    \label{eq:pair_loss}
\end{equation}
where $\sigma$ denotes the sigmoid function.

The overall post-training objective is defined as
\begin{equation}
    \mathcal{L} =
    \mathcal{L}_{\mathrm{pair}} + \mathcal{L}_{\mathrm{SDA}}.
    \label{eq:total_loss}
\end{equation}
As illustrated in Fig.~\ref{fig:overview}(d), the two objectives provide complementary supervision by preserving individual image–text correspondences through pairwise matching and aligning cross-modal semantic changes through SDA loss.

%% file: sec/4_experiments.tex
\section{Experiments}
\label{sec:experiments}

\subsection{Implementation Details}
% \paragraph{Post-training configuration.}
\noindent\textbf{Post-training configuration.}
We split VarFamily-40K into 27,999 for training, 6,000 for validation, and 6,001 for testing. We post-train SigLIP-So400M/14~\cite{zhai2023siglip} on the training split using LoRA~\cite{hu2022lora} with a rank of 8 and a scaling factor of 16. The SDA loss parameters are set to $\lambda_{\mathrm{dir}}=1$, $\lambda_{\mathrm{margin}}=0.5$, and $m=0.2$. The model is trained for five epochs via AdamW~\cite{loshchilov2019decoupled}, with a learning rate of $9.375\times10^{-7}$, a batch size of 24 image–text groups, and an input resolution of $384\times384$. To further assess the applicability of our method, SigLIP-ViT-B/16 is also post-trained with an input resolution of $224\times224$, a learning rate of $10^{-5}$, and a batch size of 256 image–text groups.

% \paragraph{VLM training.}
\noindent\textbf{VLM training.}
We adopt a two-stage pipeline comprising connector pretraining and instruction fine-tuning, using Qwen2.5-0.5B~\cite{yang2024qwen25}, OpenELM-450M-Instruct~\cite{mehta2024openelm}, and Qwen3-1.7B~\cite{yang2025qwen3} as language backbones. The visual encoder is frozen throughout both stages. For a fair comparison, we use the same 1.8M examples from LLaVA~\cite{liu2024improved} and ShareGPT4V~\cite{chen2024sharegpt4v} for connector pretraining, followed by 688K examples for instruction fine-tuning. Each stage runs for one epoch, with learning rates of $10^{-3}$ and $2 \times 10^{-5}$ and global batch sizes of 256 and 128, respectively. All comparisons on the VLM share the same training recipe. Additional training details are provided in \textbf{Appendix B}.

\subsection{Evaluation Protocols}
\label{sec:evaluation_protocols}
We evaluate our method on compositional understanding, image classification, and multimodal performance to assess fine-grained semantic discrimination and the downstream benefits of our post-training method. Details of the evaluation settings are provided in \textbf{Appendix B}.

\noindent\textbf{Compositional understanding.}
We first assess image--text matching on the test split of VarFamily-40K groups and Winoground~\cite{thrush2022winoground}, examining the model's ability to distinguish subtle semantic differences and generalize beyond our constructed data. Following Winoground~\cite{thrush2022winoground}, we report text, image, and group scores, which measure the correctness of image-to-text matching, text-to-image matching, and both jointly, respectively. Since these ranking-based metrics do not capture the magnitude of similarity differences, we additionally introduce the \textbf{m}ean \textbf{p}air \textbf{m}argin (MPM), which quantifies the average similarity advantage of correct image--text matches over mismatched alternatives across the entire evaluation set:
\begin{equation}
\begin{aligned}
\mathrm{MPM} = \frac{1}{N}\sum_{i=1}^{N}
\big[&
s(x_i^0,c_i^0)-s(x_i^0,c_i^1) \\
&+s(x_i^1,c_i^1)-s(x_i^1,c_i^0)
\big],
\end{aligned}
\label{eq:mean_pair_margin}
\end{equation}
where $N$ is the number of evaluation groups and $s(x,c)$ denotes cosine similarity between normalized image and text embeddings in the learned shared representation space.

% \paragraph{Image classification.}
\noindent\textbf{Image classification.}
We also evaluate image classification to determine whether the benefits of post-training extend to downstream recognition tasks rather than remaining limited to image--text matching. We consider three protocols with different levels of task-specific adaptation. Zero-shot classification evaluates the transfer of the adapted image--text model without task-specific training, while linear probing more directly assesses whether the adapted visual encoder provides useful features for downstream classification when kept frozen. Full fine-tuning further examines whether the benefits of post-training persist after task-specific adaptation of the visual encoder. Zero-shot evaluation covers ImageNet-R~\cite{hendrycks2021many}, CIFAR-100~\cite{krizhevsky2009learning}, Oxford-IIIT Pets~\cite{parkhi2012cats}, Food-101~\cite{bossard2014food101}, and Flowers-102~\cite{nilsback2008automated}; linear probing and full fine-tuning use the latter four datasets. We report Top-1 accuracy averaged over three random seeds for the two supervised protocols.

% \paragraph{Multimodal performance.}
\noindent\textbf{Multimodal performance.}
To assess whether the benefits of post-training extend to broader multimodal understanding, we integrate the adapted visual encoder into VLMs and keep it frozen during training. We evaluate the resulting VLMs on ten benchmarks grouped into five categories. GQA~\cite{hudson2019gqa} and RealWorldQA~\cite{xai2024realworldqa} cover visual question answering, with an emphasis on compositional reasoning and real-world spatial understanding, respectively. ScienceQA~\cite{lu2022learn} and AI2D~\cite{kembhavi2016diagram} test science and diagram reasoning, respectively. MMStar~\cite{chen2024mmstar} and MME~\cite{fu2023mme} both assess visual perception and multimodal reasoning. InfoVQA~\cite{mathew2022infographicvqa}, ChartQA~\cite{masry2022chartqa}, and OCRBench~\cite{liu2024ocrbench} evaluate understanding of infographics, charts, and text-rich images, respectively. TextCaps~\cite{sidorov2020textcaps} evaluates image captioning.

\begin{table}[t]
\centering
\captionsetup{skip=4pt}
\caption{Compositional understanding on the test split of VarFamily-40K groups and Winoground.}
\label{tab:compositional_understanding}
\scriptsize
\setlength{\tabcolsep}{2.5pt}
\renewcommand{\arraystretch}{1.15}
\begin{tabular}{cllcccc}
\toprule
Dataset & Encoder & Model
& Text (\%) $\uparrow$
& Image (\%) $\uparrow$
& Group (\%) $\uparrow$
& MPM $\uparrow$ \\
\midrule
\multirow[c]{4}{*}{\rotatebox[origin=c]{90}{VarFamily-40K}}
& \multirow{2}{*}{So400M/14}
& Base
& 62.44 & 56.29 & 44.74 & 0.048253 \\
& & Ours
& \textbf{75.67} & \textbf{72.75}
& \textbf{63.06} & \textbf{0.077483} \\
\cmidrule(lr){2-7}

& \multirow{2}{*}{B/16}
& Base
& 65.14 & 61.46 & 49.13 & 0.053687 \\
& & Ours
& \textbf{74.45} & \textbf{75.27}
& \textbf{63.84} & \textbf{0.072680} \\
\midrule
\multirow[c]{4}{*}{\rotatebox[origin=c]{90}{Winoground}}
& \multirow{2}{*}{So400M/14}
& Base
& 22.50 & 5.00 & 3.75 & 0.006855 \\
& & Ours
& \textbf{29.00} & \textbf{8.75}
& \textbf{7.25} & \textbf{0.010530} \\
\cmidrule(lr){2-7}
& \multirow{2}{*}{B/16}
& Base
& 25.00 & \textbf{8.25}
& \textbf{5.25} & 0.006980 \\
& & Ours
& \textbf{28.50} & 6.00
& 5.00 & \textbf{0.008026} \\
\bottomrule
\end{tabular}
\vspace{-1mm}
\end{table}

\subsection{Results}
\label{sec:results}

% \paragraph{Compositional understanding.}
\noindent\textbf{Compositional understanding.}
Table~\ref{tab:compositional_understanding} compares the compositional matching performance of the original pretrained SigLIP (Base) and the same model after post-training with the SDA loss (Ours) on VarFamily-40K and Winoground. On the VarFamily-40K test split, our post-training method improves the group score from 44.74\% to 63.06\% for SigLIP-So400M/14 and from 49.13\% to 63.84\% for SigLIP-ViT-B/16.
Since each group contains images and captions differing in a targeted semantic concept, these gains indicate that the adapted models more effectively distinguish correct correspondences from semantically similar mismatches introduced by the controlled edits.

On Winoground, SigLIP-So400M/14 achieves improvements across all three matching scores, with the group score increasing from 3.75\% to 7.25\%. These results suggest that the learned semantic-change representations can generalize beyond the controlled variations in VarFamily-40K. Although SigLIP-ViT-B/16 improves only in text score on Winoground, its higher MPM suggests improved average separation between correct and mismatched pairs despite limited gains in ranking-based metrics. Additional comparisons with existing fine-grained vision-language alignment methods are provided  in \textbf{Appendix C} for broader empirical context. 

\begin{figure}[t]
    \centering
    \includegraphics[width=\columnwidth]{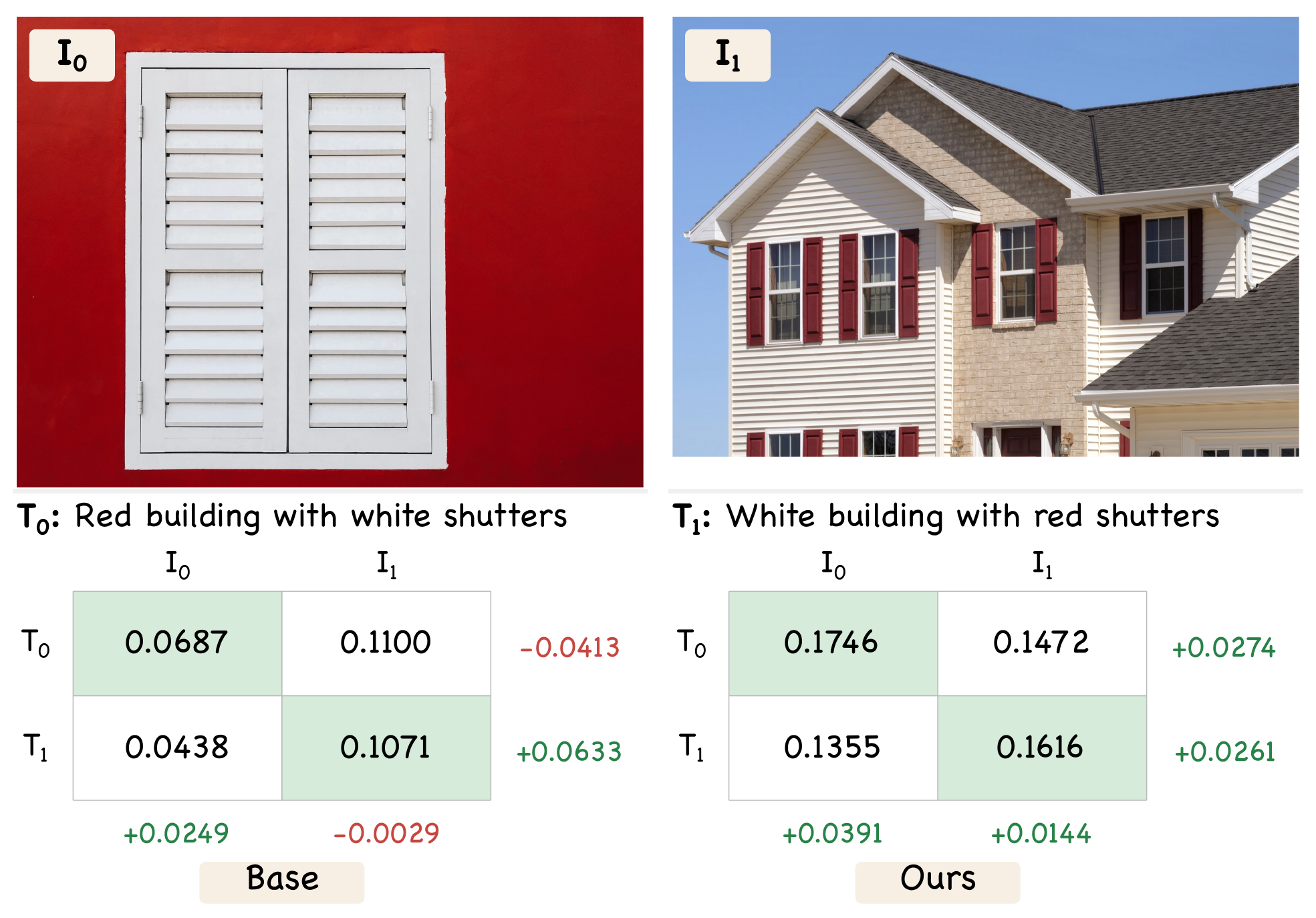}
    \caption{Case study of image--text matching on Winoground. The similarity matrices compare Base and Ours on an example involving different color–object relationships.}
    \label{fig:case_study}
    \vspace{-5mm}
\end{figure}

To further examine the improvements in fine-grained semantic understanding, we present qualitative comparisons on Winoground and VarFamily-40K. As shown in Fig.~\ref{fig:case_study}, the example contains two images with similar visual concepts but different color--object relationships. Our adapted encoder corrects the baseline's matching errors while increasing the image--text similarity for correct pairs. Moreover, this example highlights the value of MPM in capturing improvements in matching quality that remain overlooked by ranking-based metrics when the correct rankings are unchanged despite clear increases in similarity margins.

% To further illustrate the improved image–text similarity separation reflected by the higher MPM in Table~\ref{tab:compositional_understanding}, Fig.\ref{fig:case_study} compares the image–text similarity scores of the original and our adapted encoder for a representative example. The two images share similar visual concepts but differ in their color–object relationships. The original model assigns a higher similarity to the incorrect caption for the first image, whereas the SDA-adapted model correctly ranks the corresponding caption higher. Furthermore, the adapted model increases the similarity margins for both images, illustrating the improved semantic discrimination reflected by the higher MPM in Table\ref{tab:compositional_understanding}.

\begin{figure}[t]
    \centering
    \includegraphics[width=\columnwidth]{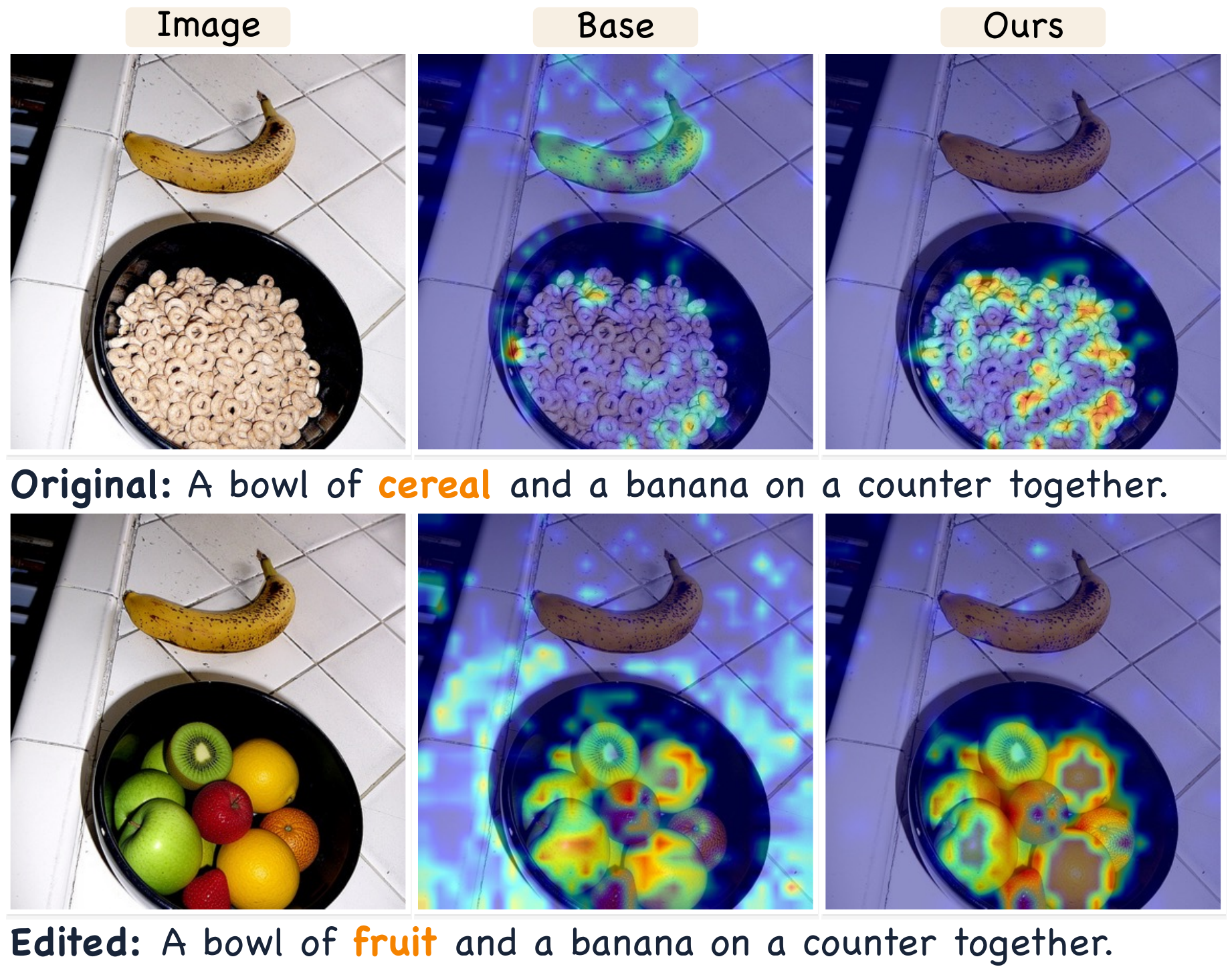}
    \captionsetup{skip=4pt}
    \caption{Visualization of gradient-weighted activation maps from the original and post-trained visual encoders, showing their spatial responses to a semantic change in the test split of VarFamily-40K.}
    \label{fig:visualization}
    \vspace{-4mm}
\end{figure}

\begin{figure*}[!t]
    \centering
    \includegraphics[width=\textwidth]{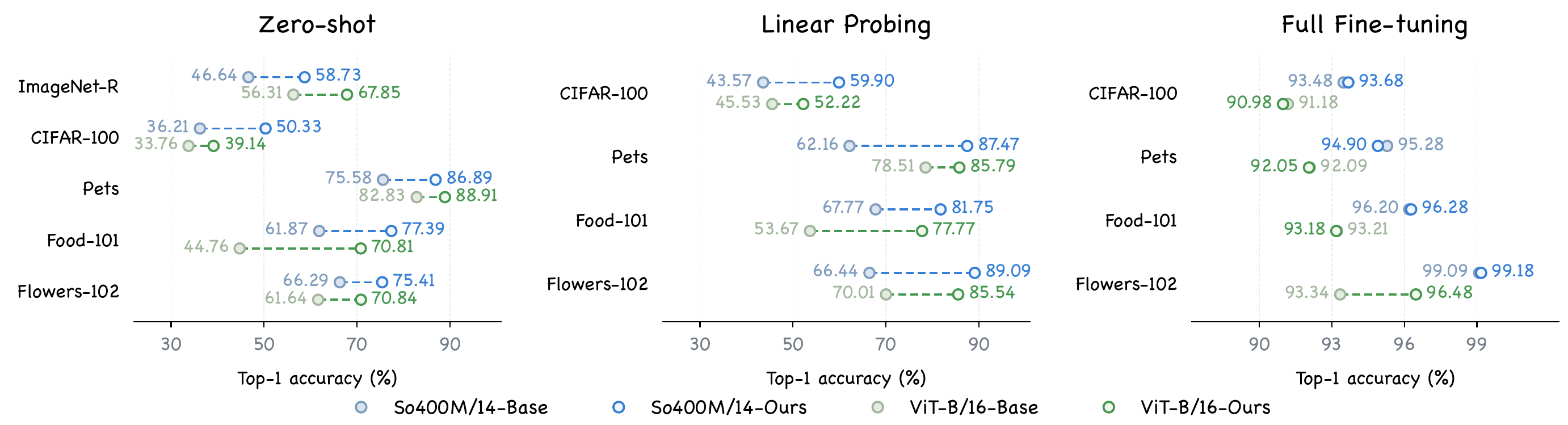}
    \captionsetup{skip=4pt}
    \caption{Image classification performance of SigLIP-So400M/14 and SigLIP-ViT-B/16 before and after post-training with the SDA loss, evaluated under zero-shot classification, linear probing, and full fine-tuning.}
    \label{fig:image_classification}
\end{figure*}

\begin{table*}[t]
    \centering
    \captionsetup{skip=4pt}
    \caption{Multimodal performance with the original and adapted SigLIP-So400M/14 visual encoders across three language backbones.}
    \label{tab:multimodal-comparison}
    \small
    \setlength{\tabcolsep}{4pt}
    \renewcommand{\arraystretch}{1.15}
    \resizebox{\textwidth}{!}{%
    \begin{tabular}{@{}*{13}{c}@{}}
        \toprule
        \textbf{Setting}
        & \makecell{\textbf{VLM Size}}
        & \textbf{Method}
        & \makecell{\textbf{GQA}\\acc$\uparrow$}
        & \makecell{\textbf{MME}\\score$\uparrow$}
        & \makecell{\textbf{ScienceQA}\\acc$\uparrow$}
        & \makecell{\textbf{InfoVQA}\\ANLS$\uparrow$}
        & \makecell{\textbf{ChartQA}\\rAcc$\uparrow$}
        & \makecell{\textbf{TextCaps}\\CIDEr$\uparrow$}
        & \makecell{\textbf{RealWorldQA}\\acc$\uparrow$}
        & \makecell{\textbf{AI2D}\\acc$\uparrow$}
        & \makecell{\textbf{MMStar}\\acc$\uparrow$}
        & \makecell{\textbf{OCRBench}\\score$\uparrow$} \\
        \midrule
        \multirow{2}{*}{Qwen2.5-0.5B}
        & \multirow{2}{*}{924M}
        & So400M
        & 58.53 & 1456.08 & 59.02 & 12.30 & 15.56
        & 0.85 & 49.93 & 48.22 & \textbf{36.33} & 300.00 \\
        & & Ours
        & \textbf{58.70} & \textbf{1467.38}
        & \textbf{60.13} & \textbf{12.79}
        & \textbf{16.68} & \textbf{0.91}
        & \textbf{50.20} & \textbf{49.09}
        & 33.73 & \textbf{306.00} \\
        \midrule
        \multirow{2}{*}{OpenELM-450M}
        & \multirow{2}{*}{890M}
        & So400M
        & 51.96 & 1184.87 & \textbf{47.49} & 11.81
        & \textbf{14.44} & \textbf{0.77} & \textbf{41.44}
        & 33.10 & 28.53 & \textbf{297.00} \\
        & & Ours
        & \textbf{52.36} & \textbf{1204.45}
        & 45.86 & \textbf{12.30}
        & 14.04 & \textbf{0.77}
        & 39.35 & \textbf{33.61}
        & \textbf{29.33} & 295.00 \\
        \midrule
        \multirow{2}{*}{Qwen3-1.7B}
        & \multirow{2}{*}{2155M}
        & So400M
        & 61.95 & 1685.48 & 70.46 & 17.25 & 18.96
        & 0.67 & 55.69 & \textbf{60.59} & 37.87 & 353.00 \\
        & & Ours
        & \textbf{62.11} & \textbf{1685.70}
        & \textbf{71.66} & \textbf{17.92}
        & \textbf{19.56} & \textbf{0.84}
        & \textbf{56.99} & 59.20
        & \textbf{38.60} & \textbf{365.00} \\
        \bottomrule
    \end{tabular}%
    }
    \vspace{-4mm}
\end{table*}
Beyond the matching case study, we visualize gradient-weighted activation maps on VarFamily-40K in Fig.~\ref{fig:visualization} to examine how the visual encoder responds to semantic changes. When the scene changes from a bowl of cereal to a bowl of fruit, our adapted encoder shows more concentrated activations in the regions corresponding to the semantic change, whereas the baseline exhibits more dispersed responses. This indicates that the adapted encoder more effectively identifies visual regions associated with the corresponding textual changes.

These results suggest that learning corresponding visual and textual changes enhances fine-grained semantic understanding, with benefits beyond the constructed training distribution. More examples are provided in \textbf{Appendix D}.

% \vspace{0.5em}
% \paragraph{Image classification.}
\noindent\textbf{Image classification.}
As shown in Fig.~\ref{fig:image_classification}, both SigLIP backbones achieve consistent improvements across all evaluated datasets under zero-shot classification and linear probing. For SigLIP-So400M/14, zero-shot accuracy on ImageNet-R increases from 46.64\% to 58.73\%, while linear-probe accuracy on Oxford Pets increases from 62.16\% to 87.47\%. Notably, the improvements under linear probing, where the visual encoder remains frozen, show that post-training enhances the transferability of visual representations rather than merely improving image--text matching.

Under full fine-tuning, the original and adapted encoders achieve similar performance, suggesting that task-specific supervision can narrow the differences between their initial representations. Overall, these results indicate that the benefits of learning corresponding semantic changes extend beyond cross-modal matching to the transferability of visual representations, particularly when the visual encoder is reused without full task-specific adaptation.

% \vspace{0.5em}
% \paragraph{Multimodal performance.}
\noindent\textbf{Multimodal performance.}
Table~\ref{tab:multimodal-comparison} presents the multimodal understanding performance of VLMs equipped with the original SigLIP-So400M/14 visual encoder and our adapted encoder across three language backbones. Replacing the original visual encoder with the post-trained encoder improves multimodal performance, particularly with Qwen2.5-0.5B~\cite{yang2024qwen25} and Qwen3-1.7B~\cite{yang2025qwen3}, where gains are observed on nine of the ten benchmarks. These improvements span visual question answering, reasoning, document and chart understanding, and text-rich visual understanding.

The visual encoder remains frozen throughout both stages of VLM training. The observed gains therefore provide evidence that our proposed SDA loss improves the downstream utility of visual representations, rather than merely enhancing image--text matching. Together, these results demonstrate the potential of semantic-change supervision to benefit multimodal understanding across different VLM configurations.

\subsection{Ablation Studies}
\label{sec:ablations}
We conduct ablation studies to examine the contributions of the visual and textual encoders, the effectiveness of individual loss components, and the influence of training data scale and edit allocation. Additional experimental analyses are provided in \textbf{Appendix C}.

% \paragraph{Visual and textual encoder contributions.}
\noindent\textbf{Visual and textual encoder contributions.}
We investigate the respective contributions of the visual and textual encoders from SigLIP-So400M/14 by evaluating different combinations of their original pretrained and post-trained versions. As shown in Table~\ref{tab:encoder_contributions}, using only the post-trained visual encoder improves the group score from 44.74\% to 62.87\% on the test split of VarFamily-40K and from 3.75\% to 8.75\% on Winoground, whereas using only the post-trained text encoder yields limited or no improvement. Moreover, using the post-trained visual encoder alone achieves comparable performance to using both post-trained encoders on VarFamily-40K and higher performance on Winoground. These results suggest that the improvements are primarily attributable to changes in the visual encoder, highlighting the effectiveness of SDA loss in improving fine-grained visual representations.

\begin{table}[t]
    \centering
    \captionsetup{skip=4pt}
    \caption{Contributions of the visual and textual encoders to compositional
    understanding. Base and Ours denote the pretrained and adapted encoders,
    respectively.}
    \label{tab:encoder_contributions}
    \small
    \setlength{\tabcolsep}{4pt}
    \renewcommand{\arraystretch}{1.15}
    \resizebox{\columnwidth}{!}{%
        \begin{tabular}{cc|cc|cc}
            \toprule
            \multicolumn{2}{c|}{Encoder} &
            \multicolumn{2}{c|}{VarFamily-40K} &
            \multicolumn{2}{c}{Winoground} \\
            \cmidrule(lr){1-2}
            \cmidrule(lr){3-4}
            \cmidrule(lr){5-6}
            Visual & Text & Group (\%) $\uparrow$ & MPM $\uparrow$
                          & Group (\%) $\uparrow$ & MPM $\uparrow$ \\
            \midrule
            Base & Base & 44.74 & 0.048253 & 3.75 & 0.006855 \\
            Ours & Base & 62.87 & 0.070876 & \textbf{8.75} & \textbf{0.011120} \\
            Base & Ours & 46.73 & 0.052966 & 3.50 & 0.006760 \\
            Ours & Ours & \textbf{63.06} & \textbf{0.077483} & 7.25 & 0.010530 \\
            \bottomrule
        \end{tabular}%
    }
    \vspace{-2mm}
\end{table}

\begin{table}[!t]
\captionsetup{skip=4pt}
\centering
\caption{Ablation study of loss components. The first row corresponds to the pretrained
SigLIP-So400M/14 encoder.}
\label{tab:loss_ablation}
\small
\setlength{\tabcolsep}{3.5pt}
\renewcommand{\arraystretch}{1.12}
\resizebox{\columnwidth}{!}{%
\begin{tabular}{ccc|ccccc}
\toprule
$\mathcal{L}_{\mathrm{pair}}$ &
$\mathcal{L}_{\mathrm{dir}}$ &
$\mathcal{L}_{\mathrm{margin}}$ &
ImageNet-R & CIFAR-100 & Pets &
Food-101 & Flowers-102 \\
\midrule
-- & -- & -- &
46.64 & 36.21 & 75.58 & 61.87 & 66.29 \\
\checkmark & & &
46.96 & 41.63 & 64.65 & 58.78 & 49.18 \\
& \checkmark & \checkmark &
57.51 & 48.83 & 86.75 & 76.22 & 74.68 \\
\checkmark & & \checkmark &
57.37 & 49.61 & 85.80 & 76.19 & 72.37 \\
\checkmark & \checkmark & &
\textbf{58.73} & 50.31 & 86.62 & 77.14 & 75.05 \\
\midrule
\checkmark & \checkmark & \checkmark &
\textbf{58.73} & \textbf{50.33} &
\textbf{86.89} & \textbf{77.39} &
\textbf{75.41} \\
\bottomrule
\end{tabular}%
}
\end{table}
\noindent\textbf{Loss components.}
Our post-training objective combines $\mathcal{L}_{\mathrm{pair}}$ with $\mathcal{L}_{\mathrm{SDA}}$, which consists of $\mathcal{L}_{\mathrm{dir}}$ and $\mathcal{L}_{\mathrm{margin}}$. To investigate their individual contributions, we evaluate different combinations of these loss components on five zero-shot image classification datasets. As shown in Table~\ref{tab:loss_ablation}, $\mathcal{L}_{\mathrm{pair}}$ alone yields inconsistent improvements, whereas $\mathcal{L}_{\mathrm{SDA}}$ alone substantially improves accuracy across all five datasets, suggesting that explicitly supervising corresponding semantic changes provides a learning signal beyond conventional image--text matching. Combining $\mathcal{L}_{\mathrm{pair}}$ and $\mathcal{L}_{\mathrm{SDA}}$ yields further improvements across all five datasets, supporting their complementary roles in learning individual image--text correspondences and cross-modal semantic changes. Removing either $\mathcal{L}_{\mathrm{dir}}$ or $\mathcal{L}_{\mathrm{margin}}$ reduces overall performance, indicating that both components contribute to the effectiveness of $\mathcal{L}_{\mathrm{SDA}}$.

\begin{table}[t]
\captionsetup{skip=4pt}
\centering
\caption{Effect of source-image diversity and edit multiplicity under a fixed budget of 10k generated samples.}
\label{tab:data_diversity}
\small
\setlength{\tabcolsep}{3.5pt}
\renewcommand{\arraystretch}{1.12}
\resizebox{\columnwidth}{!}{%
\begin{tabular}{cc|ccccc}
\toprule
Source Images & Edits per Image &
ImageNet-R & CIFAR-100 & Pets &
Food-101 & Flowers-102 \\
\midrule
10,000 & 1 &
\textbf{50.55} & \textbf{41.10} &
82.04 & \textbf{68.37} &
\textbf{70.76} \\
5,000 & 2 &
50.54 & 41.00 &
82.12 & 68.34 &
70.71 \\
2,500 & 4 &
50.54 & 41.09 &
\textbf{82.26} & 68.36 &
70.68 \\
\bottomrule
\end{tabular}%
}
\vspace{-3mm}
\end{table}

\begin{figure}[t]
    \centering
    \includegraphics[width=\columnwidth]
    {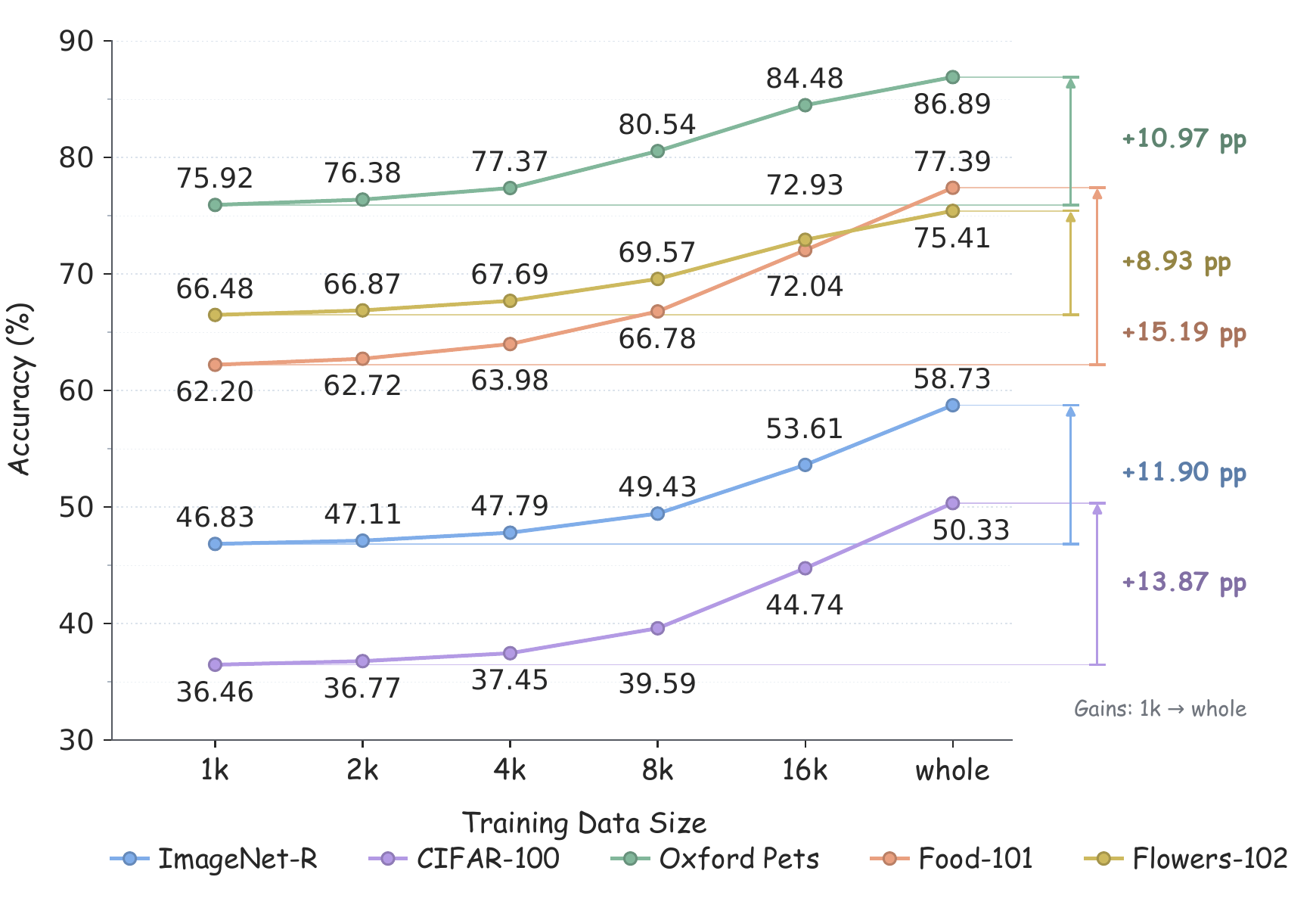}
    \caption{Effect of post-training data scale.}
    \label{fig:data_scale}
    \vspace{-4mm}
\end{figure}

% \paragraph{Training data scale.}
\noindent\textbf{Training data scale.}
Fig.~\ref{fig:data_scale} examines how the amount of training data affects our post-training method. As the number of VarFamily-40K groups increases, zero-shot classification accuracy consistently improves across all five datasets, with the full training split achieving the strongest performance. The consistent gains across all five datasets suggest that our method learns transferable visual representations from limited semantic-change supervision, with further benefits from additional training data.

% \paragraph{Edit allocation.}
\noindent\textbf{Edit allocation.}
Table~\ref{tab:data_diversity} investigates how the allocation of semantic edits across source images affects our post-training method under a fixed budget of 10k generated samples. We compare configurations ranging from 10,000 source images with one edit each to 2,500 source images with four edits each, varying source-image diversity and edit multiplicity while keeping the total number of edits unchanged. Despite these different allocations, zero-shot classification performance remains nearly identical across all five datasets, with less than 0.05 percentage points of variation in average accuracy. These results suggest that our method is relatively insensitive to how semantic edits are distributed across source images, providing flexibility in constructing semantic-change training data.

%% file: sec/5_conclusion.tex
\section{Conclusion}
\label{sec:conclusion}

In this work, we introduce Semantic Difference Alignment (SDA) loss for visual encoder post-training to improve fine-grained visual understanding. SDA loss goes beyond conventional image–text alignment by explicitly aligning corresponding semantic changes across modalities, introducing a complementary supervision signal for learning fine-grained and discriminative visual representations. To enable this learning paradigm, we constructed VarFamily-40K through targeted caption modifications, corresponding image edits, and semantic consistency filtering. Extensive experiments demonstrate improvements in compositional understanding, zero-shot classification, and linear probing across different visual backbones, alongside gains on a broad range of multimodal benchmarks when the adapted encoder is integrated into VLMs. Our findings highlight the value of learning both image–text correspondences and cross-modal correspondences between semantic changes, providing a complementary direction for improving the transferability of pretrained visual encoders.

Although our method demonstrates improvements in compositional understanding, image classification, and multimodal tasks, several limitations remain. First, VarFamily-40K relies on synthetically edited images, which may contain unintended visual changes that are not fully addressed by semantic consistency filtering. Second, our training data focuses on individual semantic changes, and the effectiveness of SDA loss when trained on more complex semantic variations remains to be evaluated. Future work will investigate more reliable data generation strategies and explore semantic-change supervision involving multiple interacting visual attributes.

%% file: sec/X_suppl.tex
\clearpage
\setcounter{page}{1}
\setcounter{section}{0}
\setcounter{table}{0}
\setcounter{figure}{0}
\setcounter{footnote}{0}

\renewcommand{\thefootnote}{\arabic{footnote}}
\renewcommand{\thesection}{\Alph{section}}
\renewcommand{\thefigure}{S\arabic{figure}}
\renewcommand{\thetable}{S\arabic{table}}
% Full-width overview
\twocolumn[{%
  \centering
  \Large
  \textbf{\thetitle}\\
  \vspace{0.5em}-- Supplementary Material -- \\
  \vspace{1.0em}
  \begin{minipage}{\textwidth}
    \normalfont\normalsize
\noindent
In this supplementary material, we describe details of VarFamily-40K and our training and evaluation procedures, and provide further evidence that our post-training method improves fine-grained visual understanding. The material is organized as follows:
\vspace{+1mm}
\begin{itemize}
    \item \textbf{Appendix}~\ref{appendix:data} provides further details on construction, filtering procedures, statistics, and splits of VarFamily-40K, complementing Sec.~3.2.
    \item \textbf{Appendix}~\ref{appendix:implementation} presents implementation details for post-training and VLM integration, along with evaluation settings for image classification, complementing Secs.~4.1 and 4.2.
    \item \textbf{Appendix}~\ref{appendix:experiments} expands on Secs.~4.3 and 4.4 with comparisons to existing methods, semantic difference retrieval, and performance analyses across semantic edit types.
    \item \textbf{Appendix}~\ref{appendix:qualitative} builds on the case studies in Sec.~4.3 by presenting further examples of image--text matching on Winoground and visual encoder responses to semantic changes in VarFamily-40K.
\end{itemize}
  \end{minipage}
  \vspace{15pt}
}]

\section{VarFamily-40K Construction and Statistics}
\label{appendix:data}
\subsection{Construction of VarFamily-40K}

\noindent\textbf{Data sources and caption rewriting.}
VarFamily-40K is constructed using images and their corresponding captions from the COCO 2017 training set~\cite{lin2014microsoft}. To introduce targeted semantic variations while preserving the original scene context, we use Qwen3.5-9B~\cite{qwen2026qwen35} to rewrite selected captions by replacing a single contiguous text span. The rewriting process follows the prompt in Table~\ref{tab:caption_rewrite_prompt}, which specifies the editing constraints and requires a JSON output containing the original span, its replacement, and the rewritten caption.

To ensure that each rewrite introduces only the intended textual change, we apply rule-based validation to the generated caption. The original span must appear exactly once in the source caption, and the rewritten caption must differ from the original one only by the specified replacement. Additional checks enforce span-length constraints and exclude comparative or relative-size expressions and unnatural phrases. If validation fails, we retry generation using the same input caption, with up to four attempts at temperatures of 0, 0.4, 0.7, and 0.9.

\begin{table*}[tbp]
\centering
\begin{tcolorbox}[
  varfamilyprompt,
  title={Qwen3.5-9B: Caption Rewriting}
]
\textbf{System message}
\begin{lstlisting}[style=varfamilyprompttext]
You create minimal visual caption edits for image editing.
Return only one JSON object with exactly these string fields:
{"original_word": "...", "new_word": "...", "new_caption": "..."}

Rules:
- Replace exactly one short contiguous span in the caption. The span may be one word or a natural phrase.
- The requested edit type will be one of: background, verb, noun, adjective, or quantity. Follow it exactly.
- background: replace one location or scene-setting phrase so the visible background changes.
- verb: replace one visually observable action phrase with a different action phrase.
- noun: replace one concrete object or subject phrase with a different common object phrase.
- adjective: replace one visually observable color, material, texture, or state phrase.
- quantity: replace one number or quantity phrase.
- Preserve every part of the caption outside the replaced span exactly, including punctuation, whitespace, capitalization, and word order.
- The replacement may have a different number of words, but keep it short: usually 1-4 words.
- Choose the span boundaries so the new caption sounds natural and common in everyday English.
- The new caption must make sense by itself. Avoid awkward or rare phrases such as "a slice of burger"; prefer common alternatives such as "a burger".
- Do not choose obscure objects, rare foods, rare locations, or unusual activities. Prefer common, visually obvious concepts.
- The new phrase must not be a synonym of the original phrase.
- Do not modify comparative or relative-size adjectives, and do not introduce them.
- Treat all size/scale words as comparative for this task. Never choose words such as small, big, large, tall, short, tiny, huge, jumbo, smaller, larger, taller, shorter, bigger, more, less, or fewer as original_word or new_word.
- Prefer diverse semantic changes. Do not default to size adjectives or quantity changes unless that edit type was explicitly requested.
- The edit must make the visible scene clearly different, remain grammatical, and be possible to render in an image.
- original_word must exactly equal the replaced span from the original caption.
- new_word must exactly equal the replacement span in the new caption.
\end{lstlisting}

\textbf{User message}
\begin{lstlisting}[style=varfamilyprompttext]
Caption: {caption}
Edit type: {edit_type}
\end{lstlisting}
\end{tcolorbox}
\caption{\textbf{Caption rewriting prompt for Qwen3.5-9B.} The system message specifies the rewriting constraints and output format, while the user message provides the original COCO caption and the requested edit type.}
\label{tab:caption_rewrite_prompt}
\vspace{-3mm}
\end{table*}

\noindent\textbf{Image editing.}
To translate textual variations into corresponding visual changes, we use FLUX.2 Klein 9B~\cite{bfl2026flux2klein} to generate edited images, conditioned on the original images and validated caption rewrites. As shown in Table~\ref{tab:image_edit_prompt}, the editing prompt specifies the original and replacement spans while instructing the model to preserve unrelated visual content. To satisfy the input requirements of FLUX.2 Klein 9B, image dimensions are adjusted to multiples of 16 while approximately preserving the original aspect ratio. The original and edited images within each group are maintained at the same resolution. We use the default inference configuration (4 denoising steps, guidance scale 1.0).
\begin{table*}[tbp]
\centering
\begin{tcolorbox}[
  varfamilyprompt,
  title={FLUX.2 Klein 9B: Image Editing}
]
\textbf{Editing prompt}
\begin{lstlisting}[style=varfamilyprompttext]
In the input image, visibly replace "{original_word}" with a clearly recognizable "{new_word}". Allow changes required for that replacement only. Preserve all unrelated content, lighting, camera angle, and style.
\end{lstlisting}
\end{tcolorbox}
\caption{\textbf{Image editing prompt for FLUX.2 Klein 9B.}
The placeholders specify the original and replacement spans generated during caption rewriting.}
\label{tab:image_edit_prompt}
\end{table*}

\noindent\textbf{Pixel-level prefiltering.}
To exclude edited images with insufficient visual changes, we apply pixel-level prefiltering to each original--edited image pair. We first use Gaussian smoothing~\cite{lindeberg1994scale} with a radius of 2 to reduce sensitivity to minor pixel-level variations. Let $D_{p,c}$ denote the absolute difference between the smoothed images at pixel $p$ and color channel $c$. We measure the magnitude and extent of visual changes using the mean absolute error (MAE)~\cite{willmott2005advantages} and the changed-pixel fraction, respectively:
\begin{equation}
\begin{aligned}
d_{\mathrm{MAE}}
&= \frac{1}{3|\Omega|}
\sum_{p\in\Omega}\sum_{c=1}^{3} D_{p,c},\\
r_{\mathrm{change}}
&= \frac{1}{|\Omega|}
\sum_{p\in\Omega}
\mathbb{I}\left[\max_c D_{p,c}\geq 20\right],
\end{aligned}
\label{eq:pixel_prefilter}
\end{equation}
where $|\Omega|$ is the total number of image pixels, and $\mathbb{I}[\cdot]$ is the indicator function. Pixel intensities are measured on a scale of 0--255. Only candidates satisfying both $d_{\mathrm{MAE}}\geq 2.0$ and $r_{\mathrm{change}}\geq 0.01$ are retained.
\begin{table*}[tbp]
\centering
\begin{tcolorbox}[
  varfamilyprompt,
  title={Qwen3-VL-8B-Instruct: Semantic Consistency Filtering}
]
\textbf{System message}
\begin{lstlisting}[style=varfamilyprompttext]
You are a strict visual quality inspector for edited image-caption pairs.
You will receive Image A (original), Image B (edited), Caption A (original),
and Caption B (edited).
Evaluate the edited pair and all four image-caption combinations independently.
Return only one JSON object with exactly these fields:
{
  "accept": true,
  "score": 5,
  "caption_a_matches_image_a": true,
  "caption_b_matches_image_b": true,
  "caption_a_matches_image_b": false,
  "caption_b_matches_image_a": false,
  "reason": "..."
}

Edited-pair quality assessment:
- Accept only if Image B clearly supports the full Caption B and the intended
  replacement is visible and correctly represented.
- Reject if the target object, action, attribute, quantity, or setting is absent,
  ambiguous, incorrect, or less consistent with Image B than the original phrase.
- Reject if major unrelated content contradicts Caption B.
- Use score 5 for clearly correct edits, 4 for minor harmless ambiguity,
  and 3 or lower for rejection.
  
Four-way consistency assessment:
- Judge each image-caption combination independently based on visible content.
- A match is true only when the image clearly supports the full caption,
  especially the changed semantic content.
- Do not infer absent or ambiguous visual details. Keep decisions logically
  consistent, including when the two images are visually indistinguishable.
- The desired pattern is (true, true, false, false), corresponding to
  (A,A), (B,B), (B,A), and (A,B), but report the actual judgments.
- Provide one concise English sentence explaining the quality verdict
  and four-way matching decisions.
\end{lstlisting}

\textbf{User message}
\begin{lstlisting}[style=varfamilyprompttext]
Caption A (original): {orig_caption}
Caption B (edited): {new_caption}
Intended replacement: {original_word} -> {new_word}
Evaluate all four combinations using Image A (original) and Image B (edited).
Return the complete JSON verdict only, including accept, score, and all four
matching decisions.
\end{lstlisting}
\end{tcolorbox}
\caption{\textbf{Semantic consistency filtering prompt for Qwen3-VL-8B-Instruct.}
The prompt specifies the edited-pair quality assessment, four-way image--text matching criteria, and JSON output format.}
\label{tab:semantic_filter_prompt}
\vspace{-8mm}
\end{table*}

\noindent\textbf{Semantic consistency filtering.}
Although pixel-level filtering identifies candidates with measurable visual changes, it cannot determine whether these changes correspond to the intended semantic edits. We therefore employ Qwen3-VL-8B-Instruct~\cite{bai2025qwen3vl} to assess each candidate group based on the original and edited images, their corresponding captions, and the intended text replacement. The model evaluates whether the edited image supports the full edited caption, with particular attention to whether the intended semantic change is correctly represented. Each candidate receives an acceptance decision and a quality score from 1 to 5 and passes this assessment only if the model returns \texttt{accept = true} with a score of at least 4.

Beyond the correctness of individual image--text pairs, VarFamily-40K requires the original and edited states to be semantically distinguishable. An edited image may support its corresponding caption while remaining compatible with the alternative caption, introducing ambiguity into semantic-change supervision. To address this, the model independently evaluates all four image--text combinations within each group. We retain only candidates with the matching pattern $(\mathrm{true},\mathrm{true},\mathrm{false},\mathrm{false})$, indicating that each image supports its corresponding caption but not the alternative caption. Both the quality assessment and four-way consistency check are performed in a single inference pass at temperature 0, using the prompt in Table~\ref{tab:semantic_filter_prompt}.

\clearpage
\subsection{Statistics of VarFamily-40K}
VarFamily-40K comprises 40,000 image--text groups, as defined in Sec.~3.1, each containing an original image--text pair and its semantically edited counterpart. This yields 80,000 image--text pair instances. While each group introduces a newly generated edited image, multiple groups may share the same original COCO image through different caption annotations. Consequently, the 40,000 groups are constructed from 34,442 unique source images, yielding 74,442 images in total, including 40,000 edited images. We further analyze the distribution of semantic variations, report filtering statistics with representative examples, and describe the dataset splits.

\begin{figure}[t]
    \centering
    \includegraphics[width=\columnwidth]{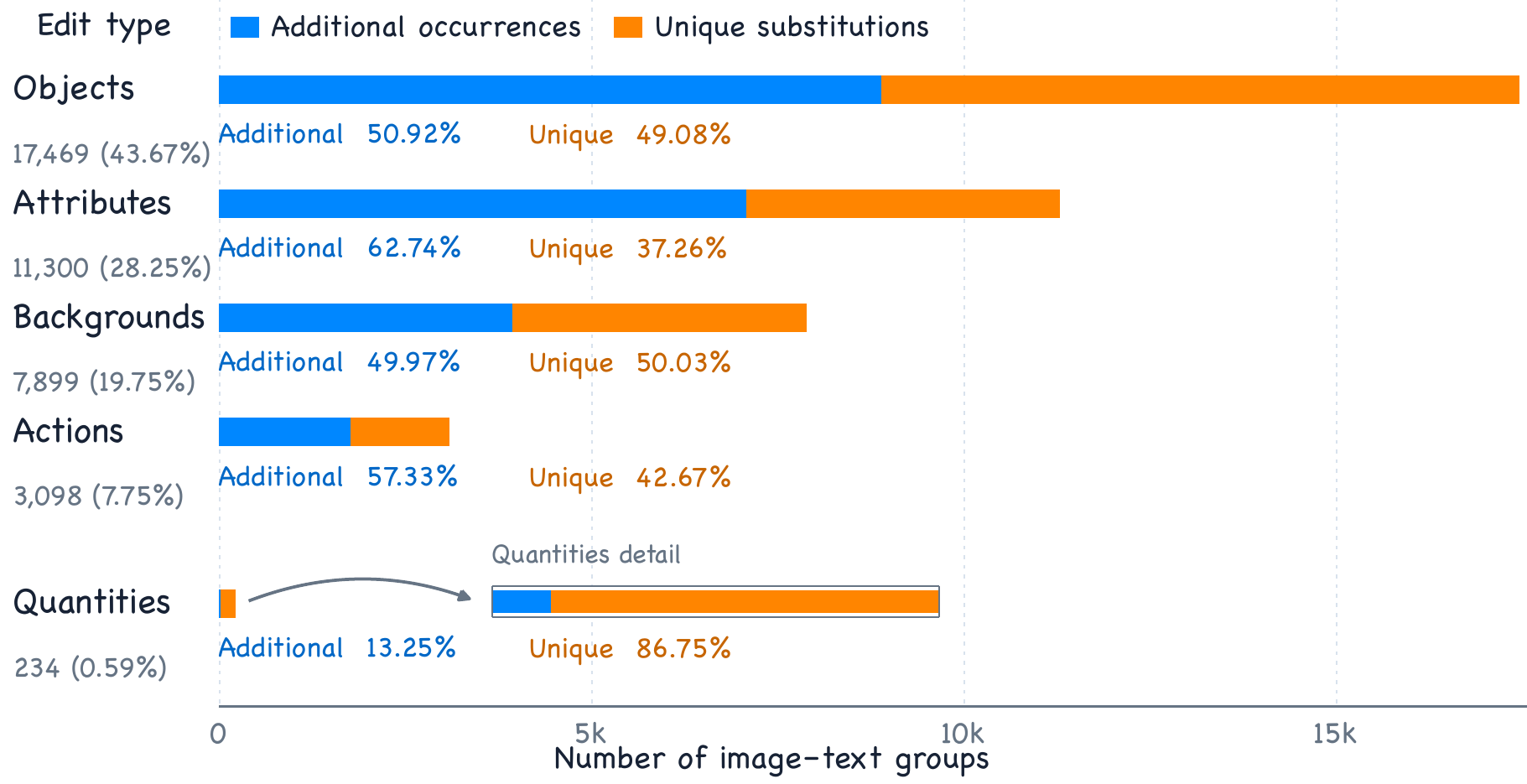}
    \caption{\textbf{Distribution of image--text groups across five edit categories.} Each bar shows the number of unique directed substitutions and additional occurrences within the category. Percentages below each category indicate its share of the dataset, while those below the bars indicate the proportions of unique substitutions and additional occurrences within that category.}
    \label{fig:semantic_variation_statistics}
\end{figure}

\noindent\textbf{Semantic variation statistics.}
We categorize the semantic variation in VarFamily-40K into five types: objects, attributes, actions, backgrounds, and quantities, corresponding to the noun, adjective, verb, background, and quantity edits in the caption rewriting prompt. As shown in Fig.~\ref{fig:semantic_variation_statistics}, object edits account for the largest share of groups (43.67\%), followed by attribute (28.25\%), background (19.75\%), action (7.75\%), and quantity edits (0.59\%). Across the 40,000 groups, we identify 17,732 unique directed text substitutions. Some substitutions recur across different source images, allowing the same semantic change to be represented in distinct visual contexts.

\begin{figure*}[t]
    \centering
    \includegraphics[width=\textwidth]{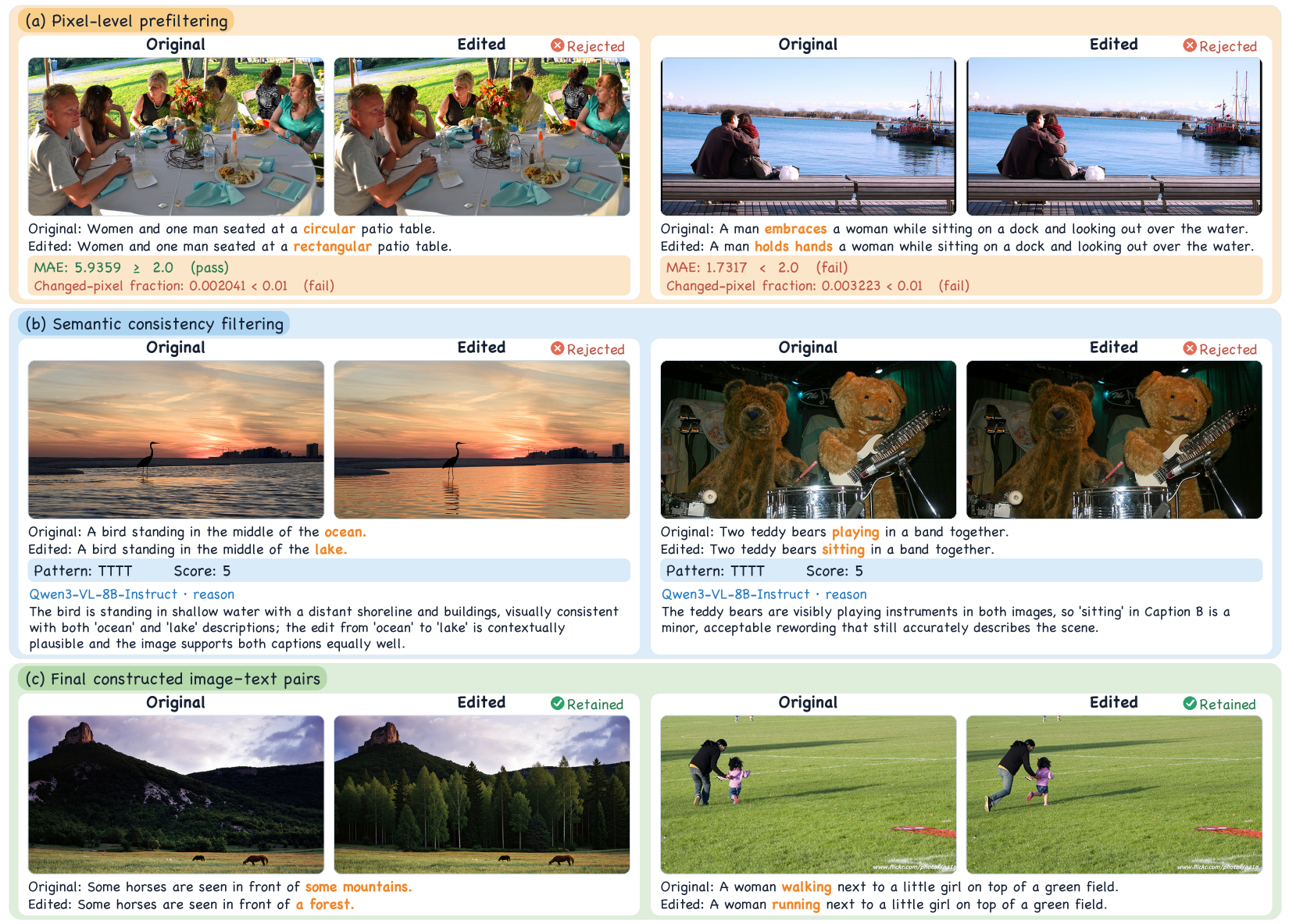}
    \caption{\textbf{Examples illustrating the two-stage filtering process of VarFamily-40K.} (a) Pixel-level prefiltering rejects candidate edits with insufficient visual changes, as measured by mean absolute error (MAE) and the fraction of changed pixels. (b) Semantic consistency filtering rejects edits that introduce ambiguous image--text correspondences, even when the edited image receives a high quality score. (c) Examples retained after both stages, where the visual changes correspond to the targeted textual substitutions and distinguish the original and edited semantic states.}
    \label{fig:filtering_cases}
\end{figure*}

\noindent\textbf{Filtering statistics and examples.}
Starting from COCO 2017 images and their rewritten captions, FLUX.2 Klein 9B generates candidate image edits, which undergo pixel-level prefiltering and semantic consistency filtering to construct VarFamily-40K. Pixel-level prefiltering retains 87.70\% of evaluated candidates, removing edits with insufficient visual changes. Of the 55,917 candidates evaluated for semantic consistency, 40,000 (71.53\%) are retained to form VarFamily-40K. Fig.~\ref{fig:filtering_cases} illustrates representative examples rejected due to insufficient visual changes or ambiguous semantic correspondences, alongside examples that satisfy both filtering criteria.

\noindent\textbf{VarFamily-40K splits.}
We partition VarFamily-40K into training, validation, and test sets at the source-image level. Since multiple image--text groups may share the same COCO image through different captions and semantic edits, we assign all groups derived from the same COCO image ID to a single split. This ensures that the held-out evaluation uses source images not included in the training split, allowing us to assess semantic-change understanding beyond the source images encountered during post-training. 

\section{Training and Evaluation Details}
\label{appendix:implementation}

\subsection{Post-training and VLM Integration}
\noindent\textbf{Post-training implementation.}
LoRA~\cite{hu2022lora} is applied to the attention output projections and both linear layers of the feed-forward networks in the visual and textual Transformer blocks. We also adapt the output projection and feed-forward layers in the visual attention pooling module. The query, key, and value projections remain frozen. In addition to the LoRA parameters, we update the logit scale and bias used in the pairwise sigmoid objective, allowing the matching logits to be calibrated as the representations change. All other pretrained parameters remain frozen.

\noindent\textbf{VLM integration.}
After post-training, we integrate the adapted visual encoder, including its learned LoRA parameters, into the VLM. The SigLIP~\cite{zhai2023siglip} text encoder is not used in the downstream VLM, where textual inputs are processed by the language backbone. Both the pretrained visual encoder weights and the LoRA parameters remain frozen during connector pretraining and instruction fine-tuning, which allows us to evaluate the downstream utility of the post-trained visual representation. The SDA loss and pairwise sigmoid objectives are used only during SigLIP post-training.

\subsection{Image Classification Settings}

\noindent\textbf{Benchmark dataset splits.}
Linear probing and full fine-tuning use identical data partitions to ensure consistent comparisons across adaptation protocols. For CIFAR-100~\cite{krizhevsky2009learning}, Food-101~\cite{bossard2014food101}, and Oxford-IIIT Pets~\cite{parkhi2012cats}, we construct stratified 90/10 training--validation splits from the official training sets while retaining the official test sets. Flowers-102~\cite{nilsback2008automated} follows its official training, validation, and test splits. Zero-shot classification is evaluated on the same official test sets as well as ImageNet-R~\cite{hendrycks2021many}.

\noindent\textbf{Zero-shot classification.}
We use multiple prompt templates to reduce sensitivity to the wording of individual class descriptions. For ImageNet-R and CIFAR-100, we adopt the 80 ImageNet prompt templates from CLIP~\cite{radford2021clip}. For Oxford-IIIT Pets, Food-101, and Flowers-102, we use five templates per dataset, including dataset-specific class descriptions and variations involving close-up, bright, and cropped photographs. For example, the template \texttt{a photo of a \{\}, a type of pet.} explicitly identifies the class as a pet. We replace \texttt{\{\}} with the class name.

\noindent\textbf{Linear probing.}
We train the classifier on $\ell_2$-normalized global image features from the frozen visual encoder, assessing how readily the learned representations support classification without task-specific updates to the encoder. The classifier is optimized with AdamW~\cite{loshchilov2019decoupled} using a learning rate of $10^{-3}$, a batch size of 1,024, and zero weight decay. We use a cosine learning-rate schedule~\cite{loshchilov2017sgdr} without warmup.

\noindent\textbf{Full fine-tuning.}
For full fine-tuning, we jointly optimize the visual encoder and the same linear classification head used for linear probing. We optimize both components using AdamW, with learning rates of $10^{-5}$ for the visual encoder and $10^{-3}$ for the classifier. We use zero weight decay and a cosine learning-rate schedule without warmup.

\section{Additional Experimental Results}
\label{appendix:experiments}

\begin{table*}[!t]
    \centering
    \caption{Comparison with additional baselines on compositional understanding and image classification. All methods use SigLIP-So400M/14. Text, image, and group scores and classification accuracies are reported as percentages. MPM denotes mean pair margin. IN-R, C100, Pets, Food, and Flowers denote ImageNet-R, CIFAR-100, Oxford-IIIT Pets, Food-101, and Flowers-102, respectively. The best result in each column is shown in \textbf{bold}, and the second-best result is \underline{underlined}.}
    \label{tab:baseline_comparison}

    \setlength{\tabcolsep}{5pt}

    % Panel A: Compositional understanding
    \begin{tabular*}{\textwidth}{@{\extracolsep{\fill}}l cccc cccc@{}}
        \toprule
        \rowcolor{panelYellow}
        \multicolumn{9}{l}{\textbf{\textit{Compositional Understanding}}} \\
        \midrule
        \multirow{2}{*}{Method}
            & \multicolumn{4}{c}{VarFamily-40K}
            & \multicolumn{4}{c}{Winoground} \\
        \cmidrule(lr){2-5}
        \cmidrule(lr){6-9}
            & Text & Image & Group & MPM
            & Text & Image & Group & MPM \\
        \midrule
        Base
            & 62.44 & 56.29 & 44.74 & 0.0483
            & \underline{22.50} & 5.00 & 3.75 & 0.0069 \\
        TripletCLIP
            & 58.66 & 45.86 & 35.18 & 0.0398
            & 9.50 & 6.00 & 1.25 & 0.0065 \\
        EQSIM
            & 60.52 & 47.21 & 37.03 & 0.0428
            & 10.50 & 6.50 & 1.50 & 0.0072 \\
        NegCLIP
            & \underline{66.51} & 55.77 & \underline{46.19} & 0.0572
            & 12.50 & 8.50 & 2.75 & 0.0097 \\
        FSC-CLIP
            & 50.26 & \underline{65.01} & 39.91 & \underline{0.0586}
            & 13.00 & \textbf{12.00} & \underline{5.25} & \underline{0.0101} \\
        \midrule
        \textbf{Ours}
            & \textbf{75.67} & \textbf{72.75} & \textbf{63.06} & \textbf{0.0775}
            & \textbf{29.00} & \underline{8.75} & \textbf{7.25} & \textbf{0.0105} \\
        \bottomrule
    \end{tabular*}

    \vspace{8pt}

    % Panel B: Image classification
    \begin{tabular*}{\textwidth}{@{\extracolsep{\fill}}l ccccc cccc@{}}
        \toprule
        \rowcolor{panelYellow}
        \multicolumn{10}{l}{\textbf{\textit{Image Classification}}} \\
        \midrule
        \multirow{2}{*}{Method}
            & \multicolumn{5}{c}{Zero-shot Classification}
            & \multicolumn{4}{c}{Linear Probing} \\
        \cmidrule(lr){2-6}
        \cmidrule(lr){7-10}
            & IN-R & C100 & Pets & Food & Flowers
            & C100 & Pets & Food & Flowers \\
        \midrule
        Base
            & 46.64 & 36.21 & \underline{75.58} & 61.87 & \underline{66.29}
            & 43.57 & 62.16 & 67.77 & 66.44 \\
        TripletCLIP
            & 45.07 & 40.04 & 62.06 & 55.68 & 46.98
            & 54.22 & 69.77 & 78.38 & 63.31 \\
        EQSIM
            & 47.03 & 41.77 & 64.84 & 58.86 & 49.36
            & 55.37 & 71.87 & 78.86 & 65.85 \\
        NegCLIP
            & \underline{55.05} & \underline{48.47} & 72.58 & \underline{68.38} & 59.03
            & \textbf{61.31} & \underline{82.80} & \textbf{82.42} & \underline{77.69} \\
        FSC-CLIP
            & 48.36 & 40.55 & 63.67 & 67.08 & 51.68
            & 55.11 & 77.38 & 79.73 & 66.21 \\
        \midrule
        \textbf{Ours}
            & \textbf{58.73} & \textbf{50.33} & \textbf{86.89} & \textbf{77.39} & \textbf{75.41}
            & \underline{59.90} & \textbf{87.47} & \underline{81.75} & \textbf{89.09} \\
        \bottomrule
    \end{tabular*}
\end{table*}

\subsection{Comparison with Existing Methods}
To further evaluate our post-training method on compositional understanding and downstream image classification, we compare it with four fine-grained vision--language alignment methods: TripletCLIP~\cite{patel2024tripletclip}, EQSIM~\cite{wang2023equivariant}, NegCLIP~\cite{yuksekgonul2023when}, and FSC-CLIP~\cite{oh2024preserving}. To ensure consistency in the underlying image--text alignment objective, we use the pairwise sigmoid loss for image--text alignment in all methods, replacing the original softmax-based contrastive loss where applicable. The results are summarized in Table~\ref{tab:baseline_comparison}.

Our method achieves the highest group scores and MPM on both VarFamily-40K and Winoground~\cite{thrush2022winoground}, with group scores of 63.06\% and 7.25\%, respectively. Existing methods improve compositional discrimination through hard-negative supervision~\cite{patel2024tripletclip,yuksekgonul2023when,oh2024preserving} or image--text similarity regularization~\cite{wang2023equivariant}, but do not explicitly align changes in visual and textual representations. In contrast, our post-training method uses paired edits to enforce the correspondence between visual and textual differences while preserving individual image--text alignment through pairwise matching. The higher group scores indicate more consistent discrimination of these closely related pairs, while the higher MPM reflects greater separation between correct and mismatched pairs. Although FSC-CLIP achieves a higher image score on Winoground, our method obtains higher group score and MPM, demonstrating the benefits of semantic difference alignment for compositional matching.

We also evaluate downstream classification on ImageNet-R~\cite{hendrycks2021many}, CIFAR-100~\cite{krizhevsky2009learning}, Oxford-IIIT Pets~\cite{parkhi2012cats}, Food-101~\cite{bossard2014food101}, and Flowers-102~\cite{nilsback2008automated}. Our method achieves the highest zero-shot accuracy on each of the five datasets. Under linear probing, it ranks first on Pets and Flowers and second on CIFAR-100 and Food-101. Since the visual encoder remains frozen under this protocol, these improvements depend on recognition information already available in the adapted visual features. These results suggest that learning corresponding semantic changes improves the transferability of visual representations, extending the benefits of our method beyond image--text matching.

\subsection{Semantic Difference Retrieval}

To examine how well the learned representations capture corresponding semantic changes, we conduct semantic difference retrieval on the VarFamily-40K test split. Unlike conventional image--text matching, which evaluates individual image--caption correspondences, this task directly compares the visual and textual representation differences induced by the same semantic edits. It therefore provides a direct evaluation of the cross-modal correspondence targeted by the SDA loss.

For each test group, we compute the visual and textual difference vectors by subtracting the original embeddings from their edited counterparts, following Sec.~3.1 of the main paper. We perform retrieval in both directions: visual-to-textual (V2T), where visual differences serve as queries to retrieve textual differences, and textual-to-visual (T2V), where the query and candidate modalities are reversed. Both directions use cosine similarity for ranking. Each query is evaluated against 100 candidates, consisting of its paired difference from the same original--edited group and 99 differences from other test groups. The correct target is the difference from the same original--edited group; semantically similar changes from other groups are treated as non-matches. Recall at $K$ (R@$K$)~\cite{karpathy2015deep} is the proportion of queries whose paired target appears among the top $K$ candidates. We report R@1, R@5, and mean reciprocal rank (MRR)~\cite{voorhees1999trec}. MRR is the average, over queries, of the reciprocal of the paired target's rank.
\begin{table}[!t]
\centering
\caption{Semantic difference retrieval on the VarFamily-40K test split. R@1 and R@5 are percentages; MRR is the mean reciprocal rank. $\Delta$ denotes the gain over Base.}
\label{tab:change_retrieval}

\setlength{\tabcolsep}{3pt}
\renewcommand{\arraystretch}{1.1}

\resizebox{\columnwidth}{!}{%
\begin{tabular}{@{}l ccc ccc@{}}
\toprule
\multirow{2}{*}{Method}
& \multicolumn{3}{c}{V2T}
& \multicolumn{3}{c}{T2V} \\
\cmidrule(lr){2-4}
\cmidrule(lr){5-7}
& R@1 & R@5 & MRR
& R@1 & R@5 & MRR \\
\midrule
Base
& 64.51 & 83.94 & 0.7334
& 60.43 & 82.38 & 0.7031 \\
\textbf{Ours}
& \textbf{76.74} & \textbf{92.06} & \textbf{0.8373}
& \textbf{71.40} & \textbf{91.42} & \textbf{0.8020} \\
\midrule
$\Delta$
& +12.23 & +8.12 & +0.1039
& +10.97 & +9.04 & +0.0989 \\
\bottomrule
\end{tabular}%
}
\end{table}

As shown in Table~\ref{tab:change_retrieval}, post-training consistently improves all metrics in both retrieval directions. V2T R@1 increases from 64.51\% to 76.74\%, while T2V R@1 increases from 60.43\% to 71.40\%. The improvements in R@5 and MRR further indicate that paired differences are ranked more favorably among competing candidates.

These improvements are consistent with the two components of the SDA loss. The directional alignment term encourages visual and textual differences induced by the same semantic edit to align in direction, thereby increasing their cosine similarity. The margin term further encourages paired differences to exhibit higher similarity than competing differences from other groups. The higher retrieval scores in both directions show that the post-trained model more reliably ranks visual and textual differences from the same edit above differences from other groups.

\begin{figure*}[!t]
\centering
\includegraphics[width=\textwidth]{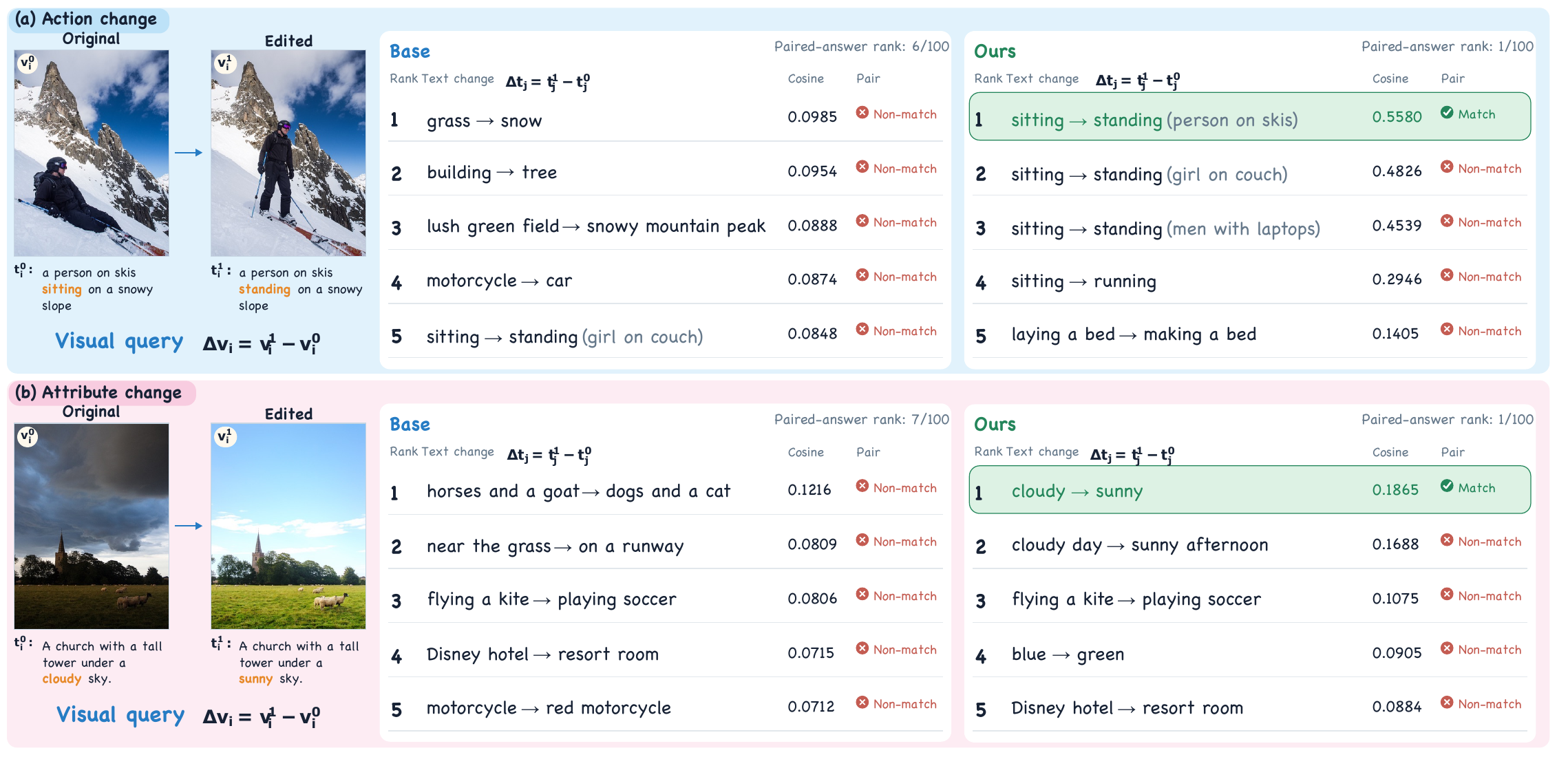}
\caption{\textbf{Visual-to-textual semantic difference retrieval.} Examples are shown for an action edit (a) and an attribute edit (b). Each visual query is computed as the embedding difference between the edited and original images. The top five textual candidates are shown with their cosine similarities, with the paired target highlighted in green and non-matches indicated by red labels. The target's rank among 100 candidates is reported above each list. Gray parenthetical descriptions distinguish the contexts in which the same textual change occurs.}
\label{fig:semantic_difference_retrieval}
\end{figure*}

Fig.~\ref{fig:semantic_difference_retrieval} illustrates two V2T retrieval examples involving action and attribute changes. Taking the action example, the visual query represents a person on skis who transitions from sitting to standing. Post-training substantially increases its similarity to textual differences expressing this action change. Among candidates describing the same action change in different contexts, the paired ``person on skis'' change receives the highest cosine similarity of 0.558, with its retrieval rank moving from sixth to first. This example illustrates that our post-training method enhances cross-modal alignment of semantic changes while preserving the ability to distinguish their specific contexts.

\subsection{Performance Across Semantic Edit Types}
To investigate the effect of post-training across different semantic edit types, we separately evaluate matching performance for objects, attributes, actions, quantities, and backgrounds in VarFamily-40K. This analysis reveals whether the overall matching improvement is consistent across different types of semantic distinctions. Table~\ref{tab:edit_type_improvements} reports the gains over Base together with the number of test groups in each category.

\begin{table}[!t]
\centering
\caption{Compositional matching gains of post-trained SigLIP-So400M/14 over Base for each semantic edit type on the VarFamily-40K test split. The Samples column gives the number of test groups per edit type. All gains are absolute differences in percentage points.}
\label{tab:edit_type_improvements}

\setlength{\tabcolsep}{5pt}
\renewcommand{\arraystretch}{1.1}

\resizebox{\columnwidth}{!}{%
\begin{tabular}{@{}l r ccc@{}}
\toprule
Edit Type & Samples
& Text $\Delta \uparrow$
& Image $\Delta \uparrow$
& Group $\Delta \uparrow$ \\
\midrule
Object
& 2,572 & +11.59 & +13.34 & +15.75 \\
Attribute
& 1,719 & +15.59 & +19.02 & +21.64 \\
Action
& 449 & +11.14 & +9.13 & +12.03 \\
Quantity
& 44 & +11.36 & +4.55 & +4.55 \\
Background
& 1,217 & +14.22 & +22.60 & +21.86 \\
\midrule
\textbf{Overall}
& \textbf{6,001}
& \textbf{+13.23}
& \textbf{+16.46}
& \textbf{+18.31} \\
\bottomrule
\end{tabular}%
}
\end{table}

Post-training improves text, image, and group scores across all five edit types, with absolute group-score gains of 15.75\%, 21.64\%, 12.03\%, 4.55\%, and 21.86\% for object, attribute, action, quantity, and background edits, respectively. Despite variations in the magnitude of improvement, the consistent gains across all categories suggest that explicitly supervising how visual and textual semantics change together enhances fine-grained semantic discrimination beyond any specific edit type.

\section{More Case Studies}
\label{appendix:qualitative}

\begin{figure*}[t]
    \centering
    \includegraphics[width=\textwidth]{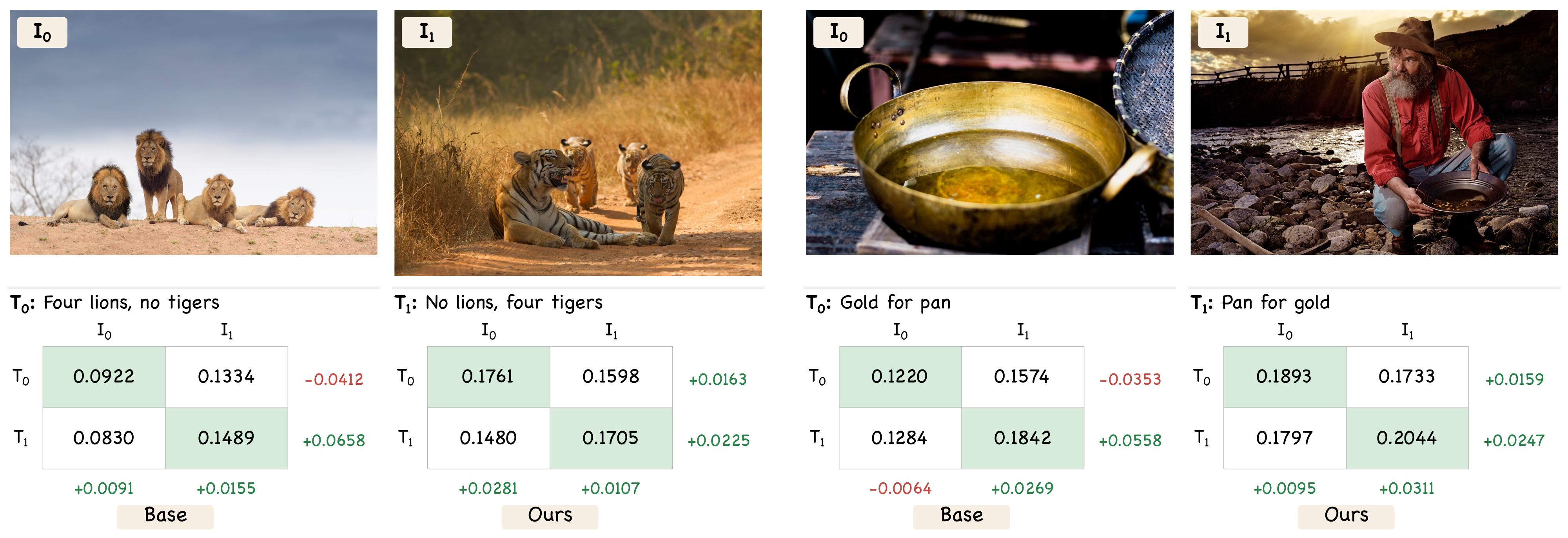}
    \caption{
    \textbf{Additional image--text matching cases on Winoground.}
    Rows and columns represent captions and images, respectively.
    Diagonal cells indicate correct image--text pairs, and the adjacent values show similarity margins between correct and incorrect matches.
    }
    \label{fig:additional_matching}
\end{figure*}

\begin{figure*}[t]
    \centering
    \includegraphics[width=\textwidth]{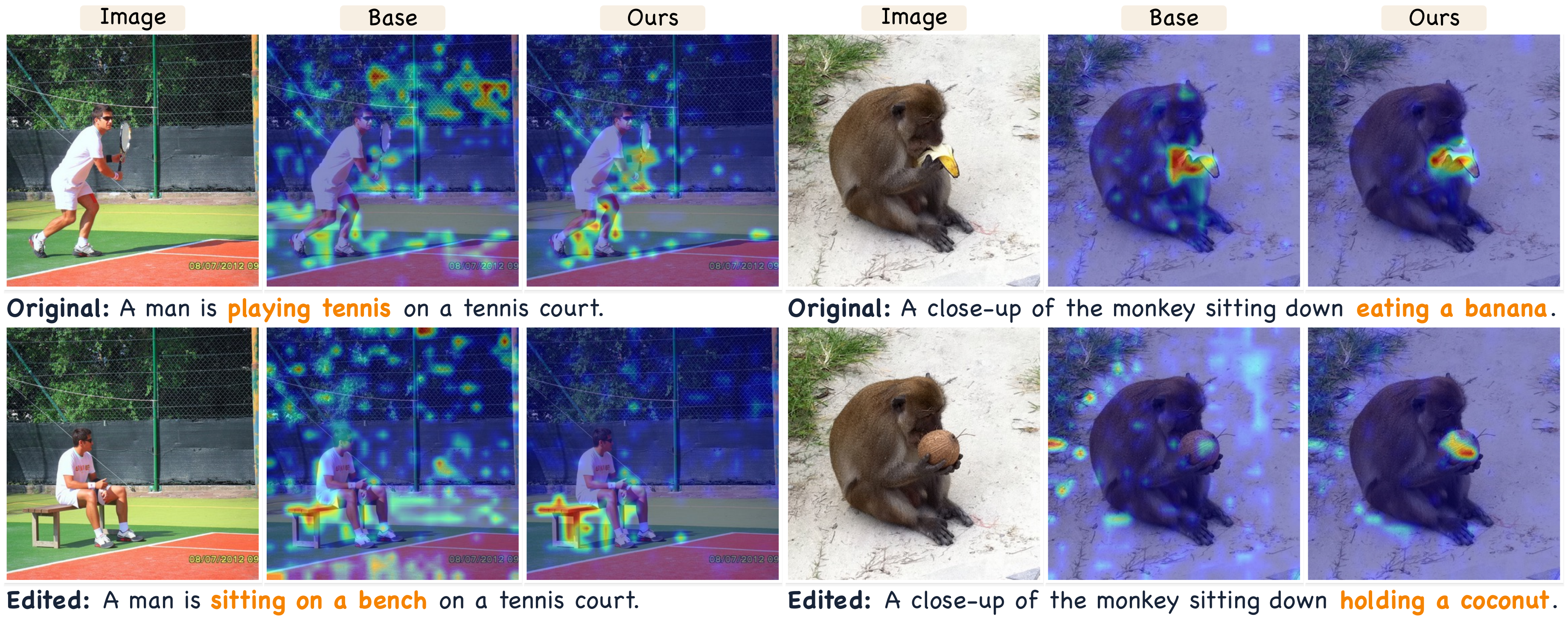}
    \caption{
    \textbf{Gradient-weighted activation maps on the VarFamily-40K test split.}
    The top and bottom rows show the original and edited examples, respectively, with modified caption spans highlighted in orange.
    }
    \label{fig:additional_activation}
\end{figure*}

Fig.~\ref{fig:additional_matching} presents two additional Winoground cases comparing the original and post-trained models.
The first contrasts ``Four lions, no tigers'' with ``No lions, four tigers''.
Both captions contain the same animal categories and count but differ in which category is present or absent.
Base incorrectly assigns the first caption a higher similarity to the tiger image, yielding a margin of $-0.0412$. 
Our model corrects this ranking, increasing the margin to $+0.0163$ while preserving the other three correct comparisons, achieving a correct group match.

The second case contrasts ``Gold for pan'' with ``Pan for gold'', distinguishing a cooking pan from the activity of panning for gold.
Although the captions contain the same words, their ordering changes the depicted meaning.
Base incorrectly associates the first caption with the gold-panning scene and misranks the captions for the cooking-pan image.
Our model resolves both errors, yielding positive margins for all four comparisons.
Together, these cases illustrate improved discrimination between captions with shared vocabulary but different semantics, suggesting that the benefits of semantic difference alignment extend beyond the controlled edits in VarFamily-40K.

Beyond image--text matching, we use gradient-weighted activation maps to examine whether the adapted visual encoder responds to the content associated with semantic changes. Fig.~\ref{fig:additional_activation} shows two examples from the VarFamily-40K test split. In the first example, ``playing tennis'' changes to ``sitting on a bench'', while the tennis-court setting remains unchanged. Base exhibits scattered activations across the fence and other background regions. After post-training, background responses are reduced, with stronger activations around the player and racket in the original image and the bench in the edited image. These responses correspond more closely to the visual content distinguishing the two activities.

The second example changes ``eating a banana'' to ``holding a coconut'', while retaining the monkey and its seated pose.
Both models exhibit activations near the banana in the original image. 
In the edited image, however, Base exhibits scattered responses across the animal and surrounding ground, whereas our model concentrates its strongest activations around the coconut and the hands holding it. These observations suggest that post-training enhances the visual encoder's sensitivity to regions associated with semantic changes within otherwise similar scenes.